\documentclass[10pt,journal,compsoc]{IEEEtran}
\newcommand{\projectname}{GRIM\xspace}

\usepackage{color}
\usepackage{ctable}
\usepackage{threeparttable}
\usepackage[normalem]{ulem}
\usepackage{float}
\usepackage{subfig}
\usepackage{enumitem}
\usepackage{longtable}
\usepackage{amsmath, amsthm, amssymb}
\usepackage{amsfonts}
\usepackage{flushend}

\usepackage{listings}
\definecolor{codegray}{RGB}{0,92,240}
\definecolor{codepurple}{rgb}{0.58,0,0.82}
\definecolor{keyword}{RGB}{186, 45, 162}
\lstdefinestyle{mystyle}{
    backgroundcolor=\color{white},   
    commentstyle=\color{codegray},
    keywordstyle=\bfseries\color{keyword},
    numberstyle=\tiny\color{codegray},
    stringstyle=\color{codepurple},
    basicstyle=\ttfamily\footnotesize,
    breakatwhitespace=false,         
    breaklines=true,                 
    captionpos=b,                    
    keepspaces=true,                 
    numbers=left,                    
    numbersep=5pt,                  
    showspaces=false,                
    showstringspaces=false,
    showtabs=false,                  
    tabsize=2,
    morekeywords={entry},
}
\definecolor{revision}{RGB}{0, 0, 0}

\usepackage{xspace}
\definecolor{xiaolong}{RGB}{181, 97, 2}
\definecolor{zhengang}{RGB}{61, 89, 171}
\definecolor{wniu}{RGB}{255, 103,104}

\ifCLASSOPTIONcompsoc
  \usepackage[nocompress]{cite}
\else
  \usepackage{cite}
\fi
\ifCLASSINFOpdf
\else
\fi
\begin{document}
%
\title{
  GRIM: A General, Real-Time Deep Learning Inference Framework for Mobile Devices based on Fine-Grained Structured Weight Sparsity
}

%
%
%
%

\author{Wei~Niu{$^\dagger$},
        Zhengang~Li{$^\dagger$},
        Xiaolong~Ma,
        Peiyan~Dong,
        Gang~Zhou,
        Xuehai~Qian,
        Xue~Lin,
        Yanzhi~Wang, 
        Bin~Ren
\IEEEcompsocitemizethanks{
\IEEEcompsocthanksitem W. Niu, G. Zhou and B. Ren are with the Department
of Computer Science, William \& Mary, Williamsburg, VA, 23185.
\IEEEcompsocthanksitem Z. Li, X. Ma, P. Dong, X. Lin and Y. Wang are with the Department of Electrical \& Computer Engineering, Northeastern University, Boston, MA, 02115.
\IEEEcompsocthanksitem X. Qian is with the Department of Computer Science, University of Southern California, Los Angeles, CA, 90089.
\IEEEcompsocthanksitem {$^\dagger$} These authors contributed equally.
}
\thanks{Manuscript received October, 2020; revised XXX, XXX.}}

%
%

\markboth{Journal of \LaTeX\ Class Files,~Vol.~14, No.~8, October~2020}%
{Shell \MakeLowercase{\textit{et al.}}: Bare Demo of IEEEtran.cls for Computer Society Journals}
%



\IEEEtitleabstractindextext{%
\begin{abstract}

It is appealing but challenging to achieve real-time deep neural network (DNN) inference on mobile devices, because even the powerful modern mobile devices are considered as ``resource-constrained'' when executing large-scale DNNs. It necessitates the sparse model inference via weight pruning, i.e., DNN weight sparsity, and it is desirable to design a new DNN weight sparsity scheme that can facilitate real-time inference on mobile devices while preserving a high sparse model accuracy. This paper designs a novel mobile inference acceleration framework GRIM that is General to both convolutional neural networks (CNNs) and recurrent neural networks (RNNs) and that achieves Real-time execution and high accuracy, leveraging fine-grained structured sparse model Inference and compiler optimizations for Mobiles. We start by proposing a new fine-grained structured sparsity scheme through the Block-based Column-Row (BCR) pruning. Based on this new fine-grained structured sparsity, our GRIM framework consists of two parts: (a) the compiler optimization and code generation for real-time mobile inference; and (b) the BCR pruning optimizations for determining pruning hyperparameters and performing weight pruning. We compare GRIM with Alibaba MNN, TVM, TensorFlow-Lite, a sparse implementation based on CSR, PatDNN, and ESE (a representative FPGA inference acceleration framework for RNNs), and achieve up to $14.08\times$ speedup.

\end{abstract}


\begin{IEEEkeywords}
Compiler optimization, Model compression, Real-time Inference, Deep neural networks, Mobile computing.
\end{IEEEkeywords}}

\maketitle

\IEEEdisplaynontitleabstractindextext

%
\IEEEpeerreviewmaketitle

\IEEEraisesectionheading{\section{Introduction}\label{sec:introduction}}
\IEEEPARstart{D}{eep} learning, as one of the most powerful machine learning techniques, has achieved extraordinary performance in computer vision and surveillance, speech recognition and natural language processing, healthcare and disease diagnosis, etc.
The two major categories of deep neural network (DNN) models are convolutional neural networks (CNNs) \cite{he2016deep,redmon2016you} and recurrent neural networks (RNNs) \cite{hochreiter1997long, cho2014properties} with unique model structures and application domains.

Due to the high efficiency and reliability, low cost, small footprint, and reprogrammability, mobile devices have been pervasively used for wireless access points, wearable electronics, robotics, autonomous driving, etc.
If equipped with deep learning, mobile devices will achieve comprehensive functionality and better performance, further enabling their broader applications \cite{bhattacharya2016smart, lane2016deepx, lane2017squeezing, ota2017deep, zhang2019deep, deng2019deep}.


It is appealing to support on-device inference by mobile devices, however, achieving efficient inference with real-time performance is still a challenging task.
Because even the powerful modern mobile devices with high-end CPUs and GPUs are considered as "resource constrained" when executing the computation-extensive and memory-hungry state-of-the-art DNN models.
Take the image classification task on ImageNet dataset \cite{deng2009imagenet} with VGG-16 model \cite{simonyan2014very} as an example.
The VGG-16 model has a size of 528 MB and sufficient learning capacity and therefore is used as one of the major pre-trained models in transfer learning \cite{torrey2010transfer}.
We executed the VGG-16 model on an embedded GPU (Adreno 640) with 16-bit floating-point for weights/intermediate results 
by using two representative mobile inference acceleration frameworks -- TensorFlow-Lite (TFLite)~\cite{TensorFlow-Lite} and TVM~\cite{chen2018tvm}.
The end-to-end inference time is 307 ms and 221 ms per frame on TFLite and TVM, respectively. However, the real-time performance typically requires 30 frames/sec i.e., 33ms/frame.

Complementary to those mobile inference acceleration approaches, DNN model compression techniques provide another possibility to efficient on-device inference.
Two main-stream model compression techniques are weight pruning \cite{han2015learning,hu2016network,wen2016learning,he2017channel,dong2017learning,li2017pruning,liu2017learning,he2018amc,liu2018rethinking,zhang2018systematic,zhang2018adam,zhu2018improving,zhuang2018discrimination,min20182pfpce,niu201926ms,liu2019autoslim,ren2019admm,zhao2019variational,yang2018efficient,ma2020pconv,dong2020rtmobile} and weight quantization \cite{courbariaux2015binaryconnect, hubara2016binarized, rastegari2016xnor, lin2016fixed, leng2018extremely}. 
Weight pruning enjoys the great flexibility of various \emph{DNN weight sparsity schemes} and has achieved very high pruning rate and accuracy. 
On the other hand, weight quantization is less supported in mobile devices especially mobile GPUs. 
Therefore, this paper leverages weight pruning as the main model compression technique, while we use 16-bit floating-point representation throughout the paper.


In this paper, we implement efficient DNN inference on mobile devices aiming for both real-time performance and high accuracy.
The difficulty of achieving real-time DNN inference on mobile devices necessitates the \emph{sparse model inference} via the weight pruning techniques, i.e., DNN weight sparsity.
However, the majority of the DNN weight sparsity schemes i.e., the non-structured sparsity and the structured sparsity are (i) incompatible with the data parallel executions on the computing systems and (ii) suffering from significant accuracy loss, respectively.

Recently, the pattern-based weight pruning techniques \cite{yang2018efficient, ma2020pconv} provide a novel weight sparsity scheme, i.e., \emph{the fine-grained structured sparsity}. 
It can be considered as enabling a certain level of flexibility in the previous (coarse-grained) structured sparsity, thus simultaneously boosting the accuracy of the structured sparsity and facilitating real-time on-device inference.
Furthermore, the work of PatDNN \cite{niu2020patdnn} leverages the pattern-based weight pruning techniques \cite{ma2020pconv} to implement fine-grained structured sparse DNN models and performs compiler optimizations to achieve real-time mobile inference. 
PatDNN is the state-of-the-art mobile inference acceleration framework.



In this paper, we design a novel mobile inference acceleration framework \underline{GRIM} that is \underline{G}eneral to both convolutional neural networks and recurrent neural networks and that achieves \underline{R}eal-time performance and high accuracy leveraging fine-grained structured sparse model \underline{I}nference and compiler optimizations for \underline{M}obiles.
We start by proposing {a new fine-grained structured sparsity scheme through the Block-based Column-Row (BCR) pruning techniques}, which works for both CNNs and RNNs.
Specifically, for any weight matrices in CNNs and RNNs, we first partition it into a number of \emph{weight blocks} and then apply independent column pruning and row pruning to each block. 
Please note that the operations in CONV layers can be transferred into the general matrix multiplication (GEMM) routine \cite{chetlur2014cudnn} and therefore we can obtain the corresponding matrix format for filters in a CONV layer.
Our BCR pruning can result in a new fine-grained structured weight sparsity that enjoys the high accuracy as the non-structured sparsity and the regularity as the course-grained structured sparsity to facilitate real-time mobile inference while overcoming their shortcomings.


Based on the new fine-grained structured sparsity scheme, our GRIM framework consists of two parts: (a) the compiler optimizations of execution code generation for real-time mobile inference; and (b) the BCR pruning optimizations for determining pruning hyperparameters and performing weight pruning.
\textcolor{revision}{
Particularly, compared with PatDNN, \projectname's new compiler optimizations depend on the newly proposed BCR pruning that is generally applicable to both CNNs and RNNs, thus requiring different designs and implementations. For example, PatDNN targets CONV computations mainly without efficient support to the fully connected (FC) layers, another kind of computation kernels in neural networks (both CNNs and RNNs). \projectname unifies both CONV and FC operations by converting CONV into GEMM (im2col), supporting CNN, RNN, and potentially the latest transformer-based models (e.g. MobileBERT and GPT-2). Correspondingly, \projectname requires a new weight compression format, a different computation reorder strategy, and a different set of tuning parameters.
Moreover, \projectname includes a more flexible/declarative Domain Specific Language (DSL) to assist finding the best-suited block configuration offline, thus saving training time. In contrast, PatDNN needs to consume more epochs to find the patterns during training.
}

We summarize the contributions of this paper as:

\vspace{-0.3em}

\begin{itemize}[leftmargin=*,noitemsep,nolistsep]
    \item We propose a new fine-grained structured sparsity scheme through the BCR pruning technique that is general to both CNNs and RNNs, achieves high accuracy for the sparse DNN model inference and facilitates real-time execution.
    
    \item We present a set of new compiler techniques to generate optimized execution codes of the sparse DNN inference implemented by the proposed BCR pruning, including a layerwise Intermediate Representation (IR) with the associated Domain Specific Language (DSL), matrix reordering, a compact model storage format, register-level load redundancy elimination, and an auto-tuning module.
    
    \item {We provide systematic optimizations of the BCR pruning. We use a decoupling strategy to reduce the pruning hyperparameter search space and our hyperparameter optimizations incorporate mobile testing with compiler optimizations.
    For performing the pruning, we formulate the BCR pruning problem and provide an ADMM-based solution.
    }
    
    \item We design the whole GRIM framework for mobile devices to realize real-time, end-to-end inference performance supporting both CNNs and RNNs. GRIM is the first unified mobile inference acceleration framework for both CNNs and RNNs.
    
\end{itemize}

For CNNs, we compare GRIM with Alibaba Mobile Neural Network (MNN) \cite{Ali-MNN}, TVM \cite{chen2018tvm}, TensorFlow-Lite (FTLite) \cite{TensorFlow-Lite}, a sparse DNN inference implementation based on CSR format \cite{greathouse2016clsparse}, and PatDNN \cite{niu2020patdnn}, across various datasets, neural network models, and CPU/GPU excutions, achieving speedups in the end-to-end inference time up to $6.84\times$, $7.09\times$, $14.08\times$, $5.81\times$, and $2.11\times$, respectively.
Since our GRIM is the first mobile inference acceleration framework for RNNs, we compare with ESE \cite{han2017ese}, a representative FPGA inference acceleration framework for RNNs.
{We achieve comparable end-to-end inference time (around 81 $us$ by GRIM and 82 $us$ by ESE) with significantly higher energy efficiency ($38\times)$ compared with ESE.}


\begin{figure}[t]
    \centering
    \includegraphics[width=0.48 \textwidth]{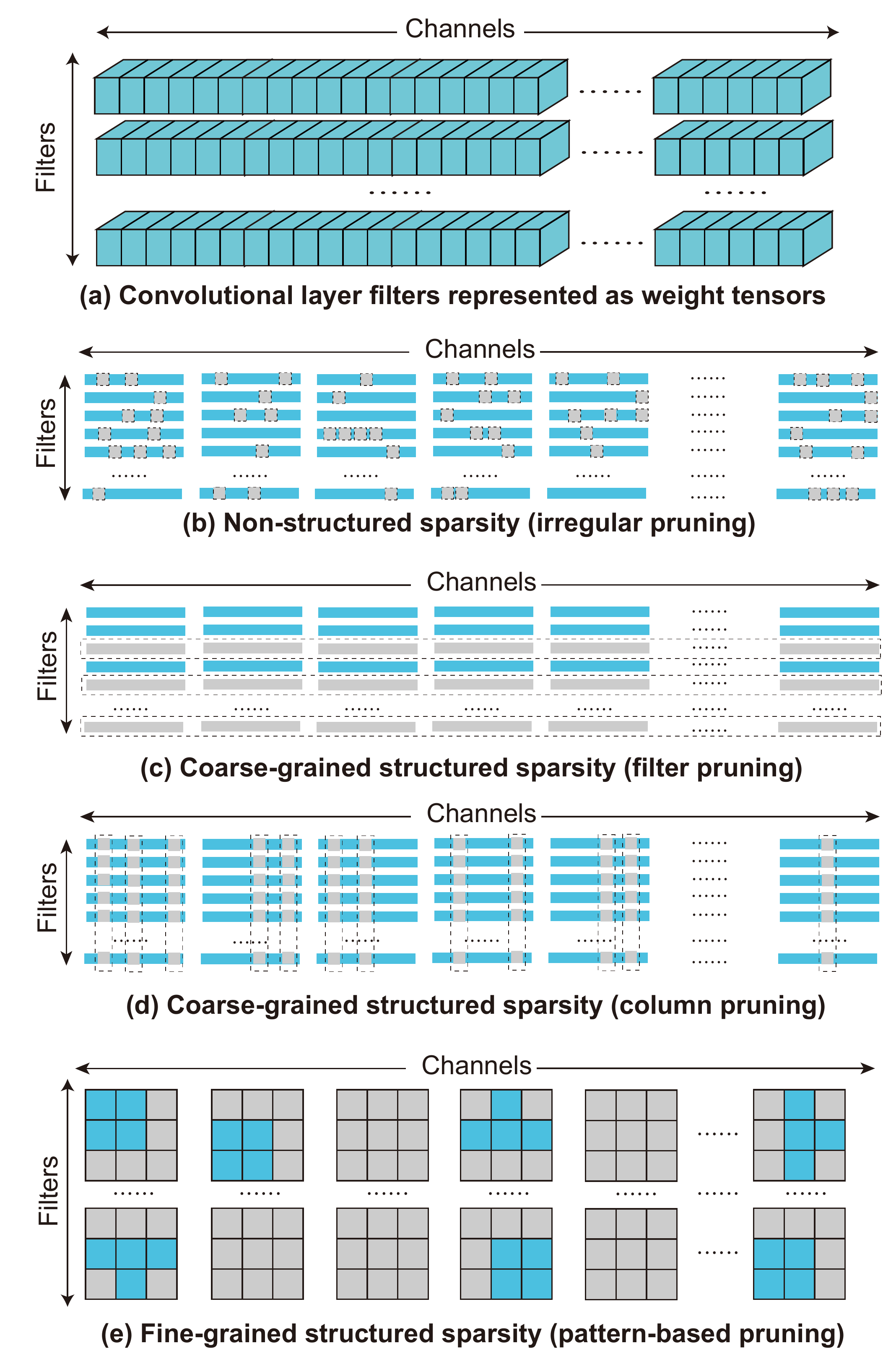}
    \caption{{\bf Weight sparsity schemes by different pruning techniques.}}
    \label{fig:sparsity_schemes}
    \vspace{-0.5em}
\end{figure}

\begin{figure*}[t]
    \centering
    \includegraphics[width=0.9 \textwidth]{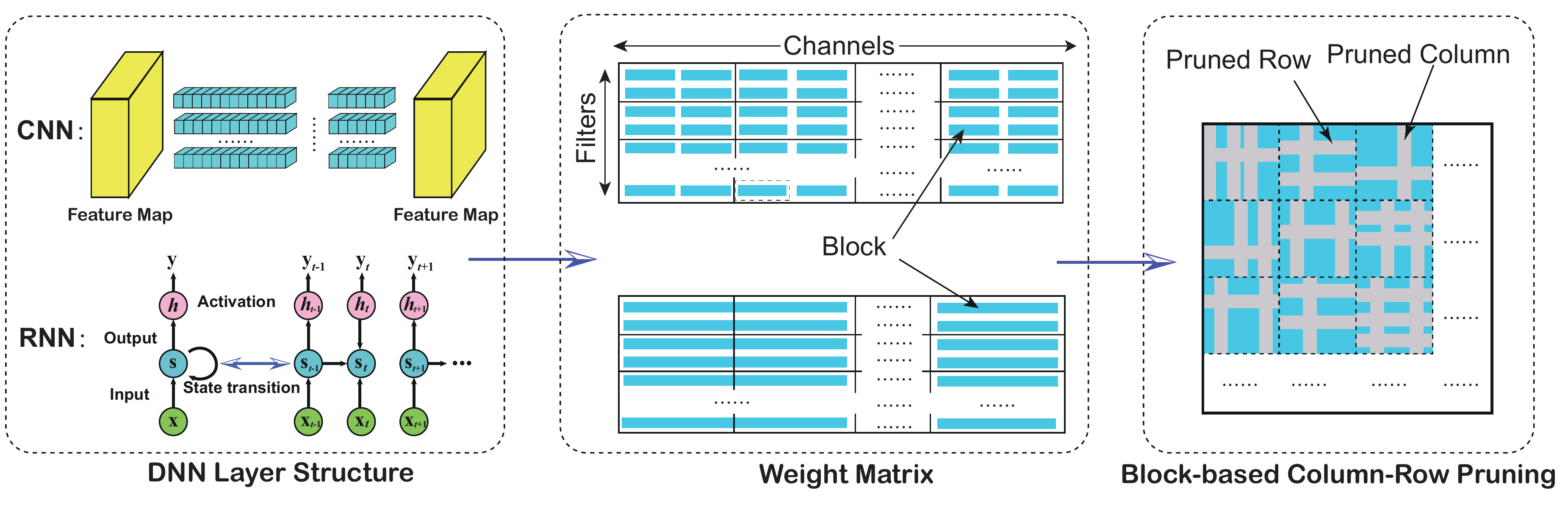}
    \vspace{-0.5em}
    \caption{{\bf Illustration of the proposed BCR pruning.}}
    \label{fig:blockpruning}
\end{figure*}

\section{Background on Sparse DNNs}

We use Figure \ref{fig:sparsity_schemes} to illustrate existing DNN weight sparsity schemes.
We use grey color to represent the pruned weights.
We start by Figure \ref{fig:sparsity_schemes} (a) showing the weight tensors in a convolutional (CONV) layer.
For Figure \ref{fig:sparsity_schemes} (b)$\sim$(d) the CONV weight tensors are transferred into the GEMM matrix format.

Figure \ref{fig:sparsity_schemes} (b) is the non-structured sparsity by the irregular weight pruning techniques \cite{dong2017learning,liu2018rethinking,zhang2018systematic}, which prune weights at arbitrary locations. The irregular pruning can achieve a high pruning rate, but the non-structured sparsity is not compatible with data-parallel executions on the computing systems. 

Figure \ref{fig:sparsity_schemes} (c) is a coarse-grained structured sparsity scheme by the filter pruning techniques \cite{wen2016learning,he2017channel,liu2017learning}  
and Figure \ref{fig:sparsity_schemes} (d) is a coarse-grained structured sparsity scheme by the column pruning techniques \cite{liu2019autoslim,zhang2018adam}. 
Filter pruning by the name prunes whole filters from a layer. (Please note that some references mention channel pruning \cite{he2017channel}, which by the name prunes some channels completely from the filters. Essentially channel pruning is equivalent to filter pruning, because if some filters are pruned in a layer, it makes the corresponding channels of next layer invalid.) Column pruning prunes weights for all filters in a layer, at the same locations.
The coarse-grained structured sparsity preserves regularity on the sparse models, but suffer from significant accuracy loss.

\begin{figure}[t]
    \centering
    \includegraphics[width=0.48 \textwidth]{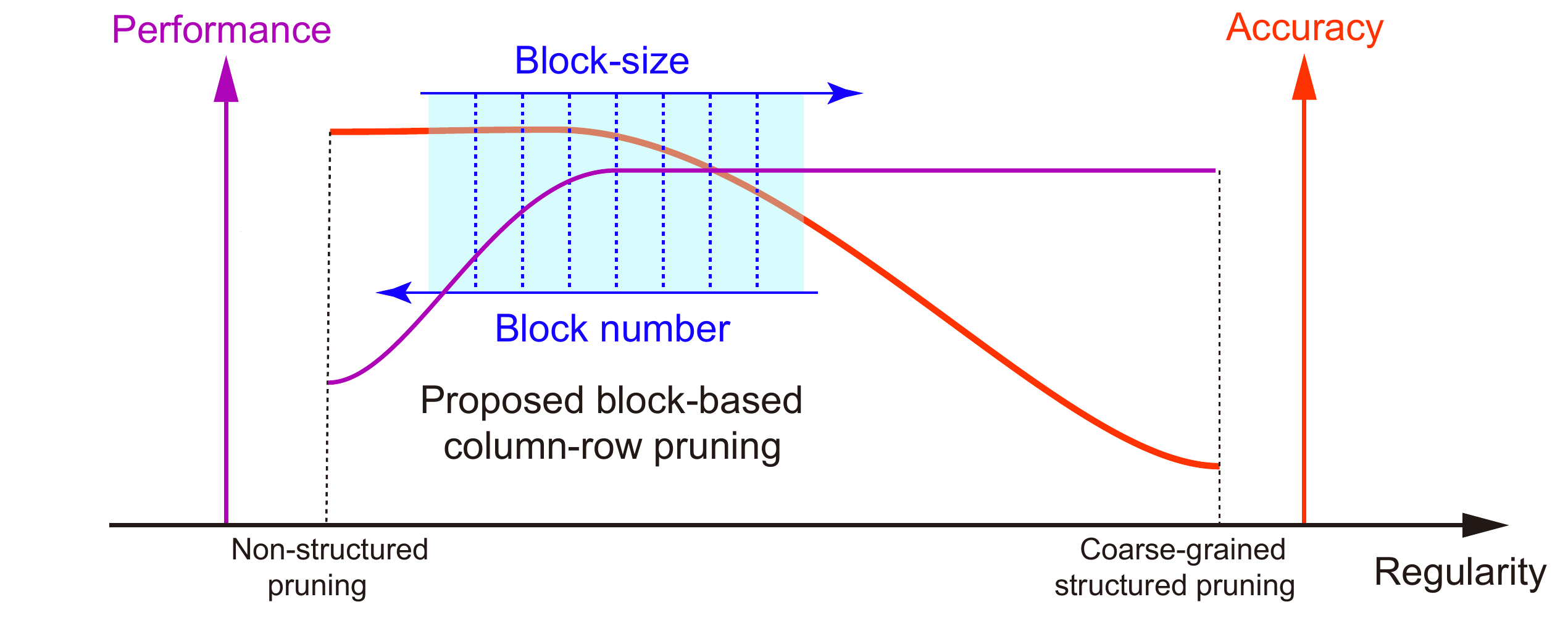}
    \caption{{\bf Relation of accuracy and performance with regularity.}}
    \label{fig:concept}
    \vspace{-0.5em}
\end{figure}

Figure \ref{fig:sparsity_schemes} (e) shows a fine-grained structured sparsity scheme by the pattern-based pruning techniques \cite{ma2020pconv,yang2018efficient,niu2020patdnn}, which are a combination of \emph{kernel pattern pruning} and \emph{connectivity pruning}.
In kernel pattern pruning, for each kernel in a filter, a fixed number of weights are pruned, and the remaining weights form specific kernel patterns. 
The example in Figure \ref{fig:sparsity_schemes} (e) is defined as 4-entry kernel pattern pruning, since every kernel reserves 4 non-zero weights out of the original 3$\times$3 kernels.
The connectivity pruning cuts the connections between some input and output channels, which is equivalent to removing corresponding kernels.
Note that the pattern-based pruning is not based on the GEMM matrix format.
More details about the differences between this fine-grained structured sparsity scheme by the pattern-based pruning and that by our BCR pruning are provided in Section \ref{sec:related}.

\section{The New Fine-Grained Structured Sparsity and \projectname Overview}

\subsection{Unified View of CNN/RNN Computation}

The layerwise computations of CNN include CONV layer computations with different kernel sizes, mostly $3\times 3$ and $1\times 1$ kernels (larger kernels such as $5\times 5$ ones may be utilized for input layer), and FC layer computations which are essentially matrix-vector multiplications. On the other hand, computations in RNNs (e.g., LSTM or GRU) are mostly FC layers (matrix-vector multiplications). It is well known that the CONV in DNNs is commonly transformed into GEMM, i.e., the multiplication of a weight matrix and an input matrix. GEMM is commonly utilized in DNN acceleration frameworks~\cite{chen2018tvm,TensorFlow-Lite}. In this way, all computation types in CNN and RNN can be unified as matrix-vector or matrix-matrix multiplication and will be treated in a unified manner through the fine-grained structured sparsity by BCR pruning.

\subsection{Motivation of Fine-Grained BCR Pruning}

\begin{figure*}[t]
    \vspace{12pt}
    \centering
    \includegraphics[width=0.95 \textwidth]{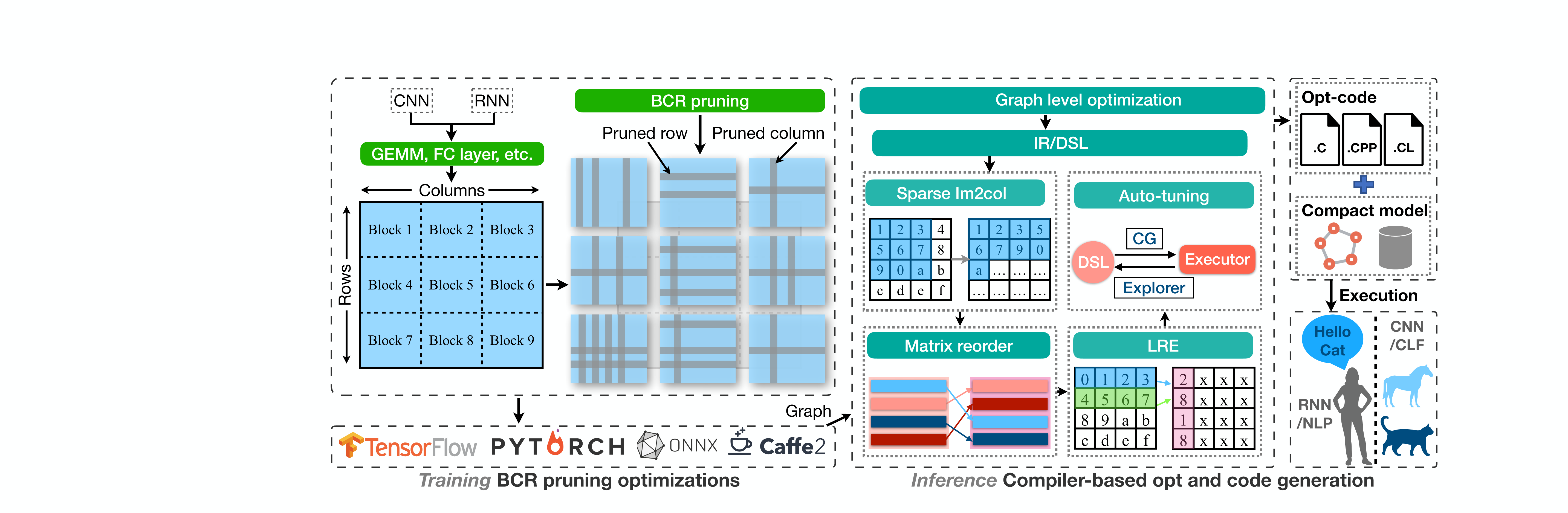}
    \caption{{\bf \projectname overview.}}
    \label{fig:sys_overview}
\end{figure*}

From a survey of recent research works, we have reached the following conclusions: (i) non-structured sparsity has the advantage of high pruning rate but is typically not compatible with the data parallel executions on the computing systems; (ii) coarse-grained structured sparsity facilitates inference acceleration but is often subject to accuracy degradation. The accuracy degradation in coarse-grained structured sparsity is especially significant for RNNs. When a whole row or column in a weight matrix (input, state-transition, or output matrix) of RNN is pruned, it assumes that a whole input or output entry is not used at all time steps. This is easy to cause intolerable accuracy loss. As a result, it is desirable to design a fine-grained structured sparsity scheme possessing more flexibility (and thus higher accuracy) while still maintaining regularity (for facilitating inference acceleration).

We propose BCR pruning to achieve this goal, which applies to different computation layers in CNNs and RNNs. For a weight matrix in GEMM or FC layer computation, we divide it into $n\times m$ blocks with equal size. We apply independent row and column pruning on each block, with potentially different pruning rates (number of pruned rows/columns) in each block, to ensure high flexibility. The remaining weights in each block still form a full matrix. The illustrative example is shown in Figure \ref{fig:blockpruning}. At first glance, BCR pruning is a tradeoff between the most flexible non-structured pruning and the most rigid structured pruning that prunes whole rows/columns. It becomes the former with block size 1-by-1 and becomes the latter with block size the same as the whole weight matrix. We will see in the following that BCR pruning is beyond a mere tradeoff, from both accuracy (pruning rate) and inference acceleration perspectives, especially with the aid of compiler optimizations.

\textbf{From the accuracy perspective}, we observe that BCR pruning obtains a significant accuracy enhancement (under the same pruning rate) compared with the most coarse-grained structured pruning that eliminates whole rows/columns, even with a small number of blocks. This is validated in various datasets under the same (ADMM-based) pruning algorithm. With a moderate 8 - 256 number of blocks in weight matrix, BCR pruning's accuracy can be similar or even surpass non-structured pruning under the same pruning rate. This is because non-structured pruning has a large search space, and it often takes too long time to converge to a desirable solution. This accuracy phenomenon is illustrated conceptually in Figure \ref{fig:concept}.

\textcolor{revision}{
Here is an example to quantitatively compare the searching space of non-structure pruning and BCR pruning. Consider pruning a layer with 64 filters and 64 channels with a $2\times$ pruning rate. The total number of possible combinations of non-structured pruning positions is C(4096, 2048), extremely large. In contrast, if using BCR pruning with a  block size of $4\times16$ and only considering the column pruning, the total number of combinations is C(1024, 512). If using filter pruning, the number of combinations is C(64, 32). The above comparison shows that the total number of BCR pruning candidate combinations is within a proper range, neither too large like non-structured pruning, nor too small like coarse-grained structured pruning. 
Although Reinforcement Learning can somehow alleviate the impact of large search space for non-structured pruning, the sampling action in DRL is generated in a randomized manner, and it is challenging to make satisfied decisions for high pruning rates. Thus, we claim that controlling the search space for weight pruning is still necessary. 
}

\textbf{From the inference acceleration perspective}, with a moderate 8 - 256 number of blocks in weight matrix, the inference acceleration performance on a mobile device can be close to the coarse-grained structured pruning, far better than the non-structured one. The most important reason is that the remaining parallelism in each block (after pruning) is still much higher than that in a mobile CPU/GPU. Taking a $1024\times 1024$ weight matrix as an example. Suppose 64 blocks are utilized and a further 8$\times$ BCR pruning rate is adopted, the average number of remaining weights per block is 2,048. These 2,048 weights form a weight matrix that is still large enough for parallelization on mobile CPU/GPU. Moreover, the overhead in column/row index storage, input and output transition, etc. can be effectively reduced through code optimization capability of compiler, and load balancing can be maintained. As a result, with the help of compiler, the inference performance can be guaranteed under fine-grained BCR pruning.

In summary, the conceptual Figure \ref{fig:concept} shows that BCR pruning is ``beyond a mere tradeoff'' of non-structured and the most coarse-grained structured pruning. Rather, it can achieve the best of both schemes, i.e., both high accuracy (pruning rate) and high inference acceleration performance, under a compiler-assisted acceleration framework.

\vspace{-0.2em}
\subsection{Overview of the \projectname Framework}

Figure~\ref{fig:sys_overview} illustrates the overview of our end-to-end \projectname acceleration framework, which consists of two major parts: (1) an execution code generation stage with the compiler-based optimizations enabled by our BCR pruning (Section~\ref{sec:inference}). This part assists inference acceleration with a given BCR pruned DNN (CNN or RNN) model and is performed offline; and (2) an optimization framework to determine the block size (for each layer) and other hyperparameters, and perform BCR pruning accordingly (Section~\ref{sec:BCRpruning}). This part is performed during the training phase.


At the high-level, \projectname represents the DNN models as computational graphs with a set of associated optimizations like TVM~\cite{chen2018tvm}. Based on this optimized baseline and by leveraging our BCR pruning, this work focuses on proposing a layerwise Intermediate Representation (and a Domain Specific Language) for each DNN layer, and designing multiple optimization and code generation techniques. Our proposed optimizations include an efficient CONV to matrix multiplication transformation (i.e., {\tt Im2col} for CNN only),  matrix reordering, a compact model storage format, register-level load redundancy elimination, and an optimized auto-tuning. These optimizations are general, applicable for both CNNs and RNNs (and associated computation types), working for both CPUs and GPUs on mobile devices. The optimized RNN and CNN models with BCR pruning can be used for various real-time workloads like natural language processing, computer vision, and video processing.

\vspace{-0.3em}
\section{Compiler Optimizations}\label{sec:inference}


\begin{figure*}[t]
    \centering
    \includegraphics[width=0.9 \textwidth]{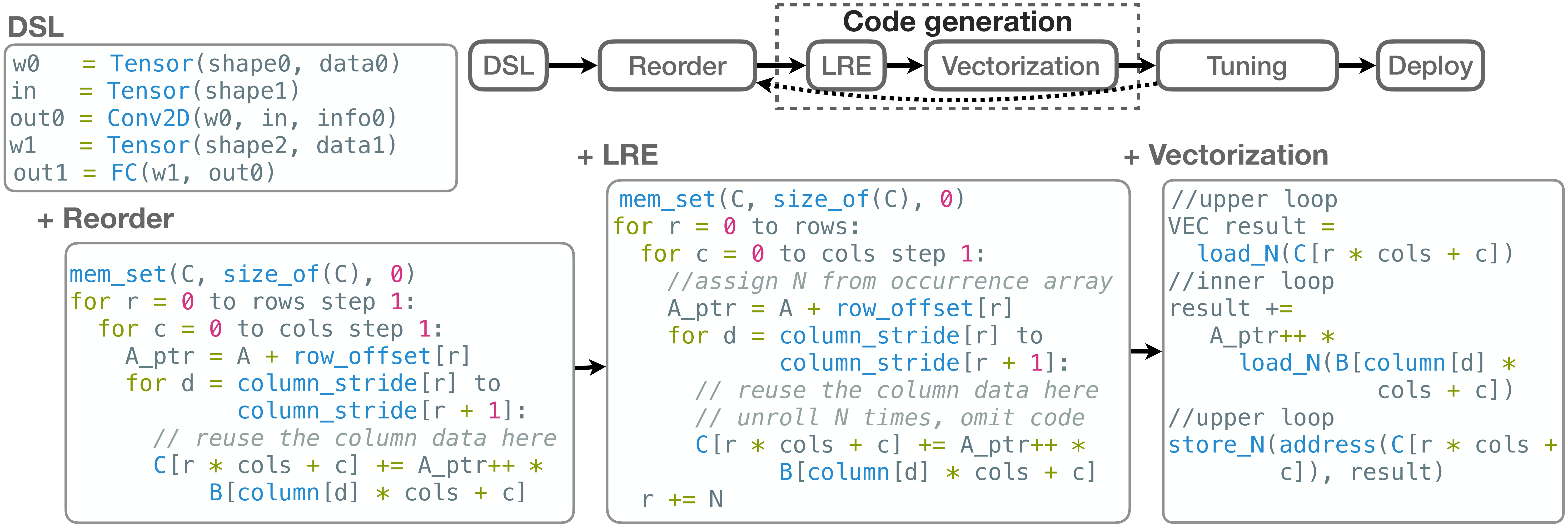}
    \caption{{\bf \projectname's compiler-based optimization and code generation flow:} compiler takes both {\tt DSL} and layerwise IR (as an example in Figure~\ref{fig:layer_ir}) to generate low-level C/C++ and OpenCL. This low-level code is further optimized with matrix reordering and our BCRC compact model storage ({\tt +Reorder}), the register-level load redundancy elimination ({\tt +LRE}), and other optimizations like vectorization ({\tt +Vectorization}). Finally, the code is further tuned by the auto-tuning module and deployed on mobile devices. 
    }
    \label{fig:compiler_controlflow}
\end{figure*}



\projectname employs a compiler-based framework to generate optimized DNN inference code on mobile devices. At the high-level, this framework represents DNN models as computational graphs like TVM~\cite{chen2018tvm} with all optimizations summarized in Table~\ref{tab:dnn-frameworks}. Based on this optimized baseline, this section focuses on the optimizations performed on each DNN layer and enabled by BCR pruning only. 

It is worth noting that although \projectname shares similar optimization objectives with PatDNN (i.e., addressing performance challenges in pruned DNN executions: {\em thread divergence} and {\em load imbalance} among threads, {\em redundant memory access}, and {\em unnecessary zero storage}), its new optimization techniques depend on the new BCR pruning that is generally applicable to both CNNs and RNNs, thus requiring very different designs and implementations comparing to PatDNN. Moreover, \projectname introduces a new DSL to improve DNN programming productivity.
Figure~\ref{fig:compiler_controlflow} shows an overview and a simplified code transformation and generation example of the \projectname compiler.

\subsection{DSL and Compiler-based Framework}

DNN models contain layers with varied computations, such as CONV, FC, pooling, etc. \projectname offers a high-level Domain Specific Language (DSL) to specify the functionality (e.g., CONV or FC), input (e.g., model, image, and intermediate results), output (e.g., intermediate and final results), and a layerwise Intermediate Representation (IR) with BCR pruning information. The input and output are in the form of tensors with different shapes. \projectname's DSL also provides a {\tt Tensor} function for users to create matrices (or tensors).

Essentially, this DSL is equivalent to the computational graph (i.e., DSL is another high-level set of functions to model the data-flow of DNN models) and they can convert to each other conveniently.  
DSL offers users the flexibility of using existing DNNs or creating new DNNs, improving the programmability (or productivity) in DNN programming. 
If a DNN already exists, \projectname transforms it into an optimized computational graph and translates this graph to DSL. Otherwise, the user writes the model code in our DSL, translates it back to a computational graph, performs high-level optimizations, and re-generates the optimized DSL code.

Figure~\ref{fig:compiler_controlflow} shows a DSL example with two connected layers, {\tt Conv2D} and {\tt FC}. {\tt Conv2D} takes a model tensor ({\tt w0}) with the shape of {\tt shape0} and data of {\tt data0} and an input feature map ({\tt in}) with the shape of {\tt shape1}, and generates a result tensor ({\tt out0}). Next, {\tt FC} takes a model tensor ({\tt w1}) with the shape of {\tt shape2} and data of {\tt data1} and previous {\tt Conv2D} output, and generates a new result tensor ({\tt out1}). 

\projectname compiler translates DSL to low-level C++ (on CPU) and OpenCL code (on GPU) and optimizes the low-level code with a set of BCR pruning enabled optimizations, such as matrix reorder, compact data storage, load redundancy elimination, configuration parameters auto-tuning, and vectorization (as Figure~\ref{fig:compiler_controlflow}). The generated code is deployed on mobile devices. 

\begin{figure}[t]
    \centering
    \includegraphics[width=0.48 \textwidth]{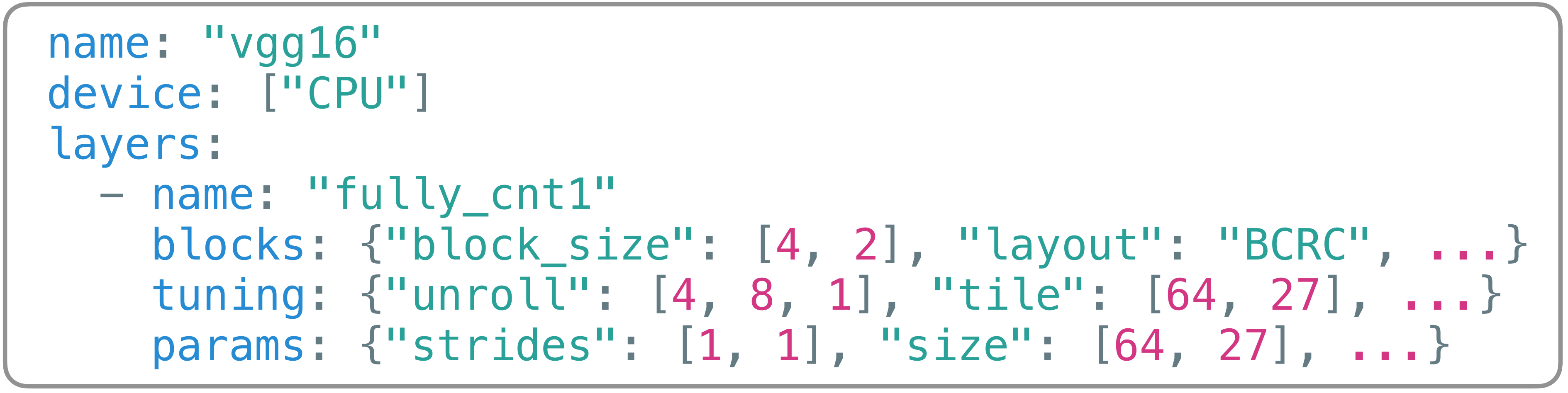}
    \caption{{\bf A layerwise IR example.}}
    \label{fig:layer_ir}
\end{figure}

\noindent{\bf Layerwise IR:} The key design of our DSL is {\em prune-aware}. It allows integrating BCR pruning information to the kernel computation by a layerwise IR (e.g., {\tt info} in the DSL example in Figure~\ref{fig:compiler_controlflow}). 
This IR provides the compiler necessary information to perform the subsequent BCR pruning-based code optimization. 
Figure~\ref{fig:layer_ir} shows more details of this IR. 
It is an {\tt FC} layer ({\tt full\_cnt1}) from {\tt vgg16}, and this IR is for {\tt CPU} optimization. It mainly consists of three aspects of information: block information (e.g., {\tt block\_size} and {\tt layout}), tuning information (e.g., {\tt unroll} factor, and {\tt tiling} size), and other basic information (e.g., {\tt strides}).
This design is general, potential to support more advanced pruning and to represent other sparsity information for further performance optimization. 

\subsection{Matrix Reordering}

\begin{figure}[t]
    \centering
    \includegraphics[width=0.475 \textwidth]{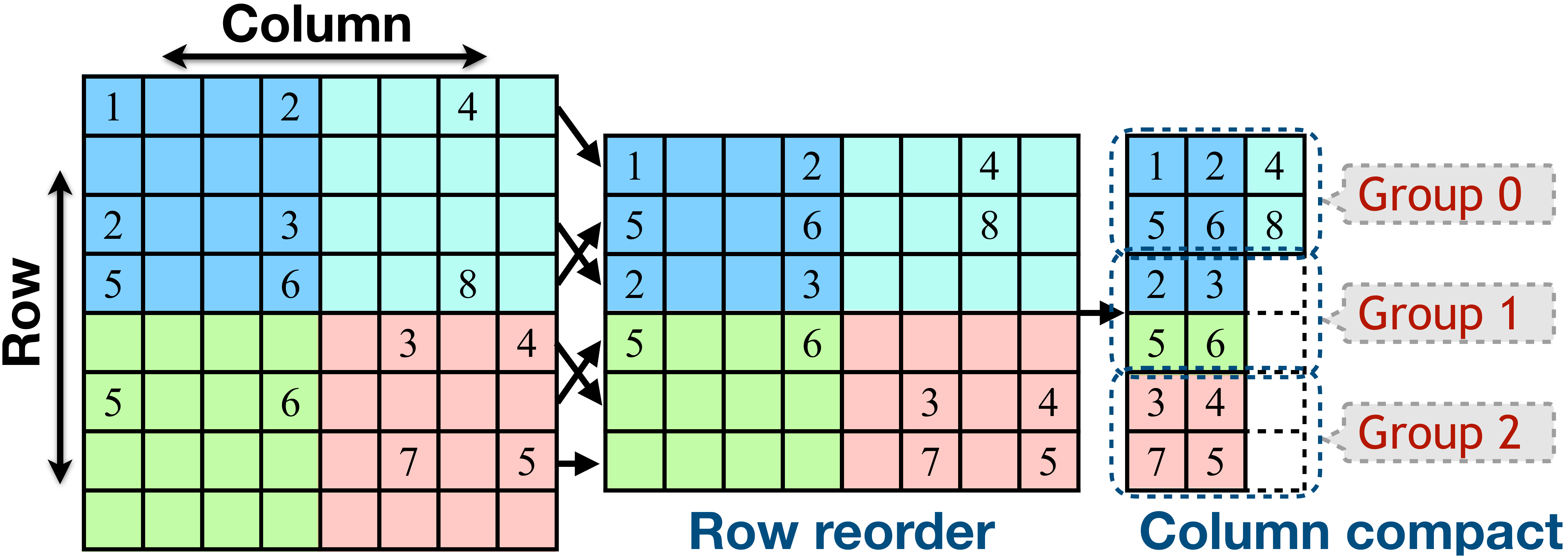}
    \caption{{\bf Matrix reordering.}}
    \label{fig:filter_reorder}
\end{figure}

BCR pruning partitions the weight matrix of a whole layer into blocks with different pruning configurations. 
Without any further optimization, it will encounter the well-known challenges for sparse matrix multiplications, i.e., heavy control-flows within each thread, load imbalance among multiple threads, and irregular memory access. Although there are many existing efforts on sparse matrix multiplications~\cite{greathouse2016clsparse,liu2015sparse}, they cannot leverage the optimization opportunities offered by BCR pruning. 
   
To address this issue, we propose a matrix reorder method based on BCR pruning. Our later evaluation demonstrates that this kind of compression and acceleration {\em co-design} significantly outperforms existing general sparse matrix multiplication optimizations that do not take the pruning characteristic into account.

Figure~\ref{fig:filter_reorder} illustrates the basic idea of matrix reorder. Because BCR pruning removes certain whole columns and rows of weights 
within a block, the remaining weights only appear in other rows and columns with a certain degree of regularity. Based on this insight, matrix reorder first reorders the rows (e.g., filters in CNN) by arranging the ones with the same or similar patterns together. Next, it compacts the weights in the column direction (e.g., kernels in CNN). At last, the rows with the same or similar computations are grouped together. 

Figure~\ref{fig:filter_reorder} shows a simplified example with only three groups and two rows in each group. Real CNN and RNN models usually have tens of groups with hundreds of rows in each group. Each group is processed by all threads in parallel, and each thread is in charge of multiple continuous rows. Thus, the computation divergence among these threads is significantly reduced.   

\subsection{Compact Model Storage (BCRC)}

After the matrix reorder, \projectname stores the model in a compact format by leveraging the BCR pruning, called BCRC (i.e. Blocked Column-Row Compact) format. BCRC aims to avoid zero-weights storage as CSR with an even better compression ratio by adopting a hierarchical index structure to remove redundant column indices generated by BCR pruning. BCRC helps to save the scarce memory-bandwidth of mobile devices. 

\begin{figure}[t]
    \vspace{12pt}
    \centering
    \includegraphics[width=0.4 \textwidth]{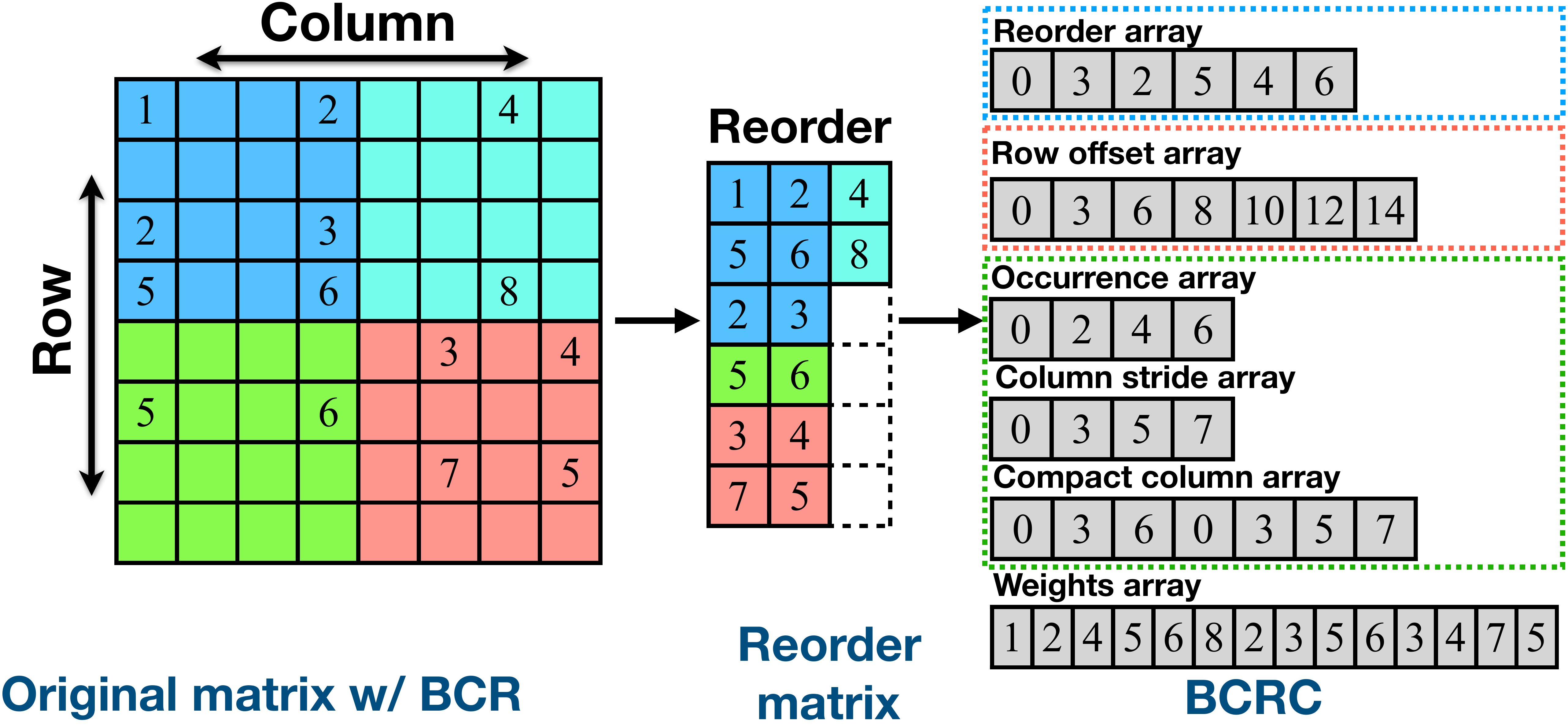}
    \caption{{\bf BCRC compact storage.}}
    \label{fig:compact_storage}
    \vspace{-0.5em}
\end{figure}

Figure~\ref{fig:compact_storage} shows a simplified example of BCRC. The original matrix with BCR pruning (left-hand side) is transformed into a compact matrix by reordering (middle) and then stored in BCRC (right-hand side). BCRC consists of six arrays: {\tt reorder}, {\tt row offset}, {\tt occurrence}, {\tt column stride}, {\tt compact column} and {\tt weights}:

\begin{itemize}[leftmargin=*,noitemsep,nolistsep]

\item {\tt Reorder array} denotes a mapping between the row id in the original matrix and the one in the reordered matrix. 
For example, the number $0$ and $3$ (in reorder array[0] and [1]) denote that the $row_0$ and $row_3$ in the original matrix are placed in the $0$ and $1$ rows, respectively, after the reorder.
\item {\tt Row offset array} denotes the offset of each row when the reordered matrix is linearized into a 1-d array (i.e. {\tt weights array}). For example, the $0$ and $3$ (in row offset array[0] and [1]) mean that the $row_0$ and $row_1$ in the reordered matrix start from index $0$ and $3$, respectively, in the 1-d {\tt weights array}.
\item The key advantage of BCRC over CSR is to use a more compact way to store the column index based on the observation that {\em multiple rows may share the same column index due to the BCR pruning}. It uses three arrays to achieve this: {\tt occurrence}, {\tt column stride} and {\tt compact column}. Here is the basic idea. {\tt Compact column array} stores the column index of each row in the reordered matrix. The {\tt column stride array} denotes the offset of the column index in each row. For example, the $0$ and $3$ (in column stride array[0] and [1]) mean that the first row in reordered matrix has the column index [$0$, $3$, $6$] (i.e. from compact column array [0] to [$2 (i.e., 3 - 1)$]).   
If two rows share the same column index, {\tt compact column array} only stores once. The {\tt occurrence array} is used to specify which rows have the same column index. For example, the first two numbers [0, 2] (in occurrence array [0] and [1]) show $row_0$ and $row_1$ have the same column index [$0$, $3$, $6$].  
\item {\tt Weights array} is to store the matrix weights in a linearized 1-d array.
\end{itemize}

The low-level code starts to support computations on BCRC from {\tt +Reorder} in Figure~\ref{fig:compiler_controlflow}.


\subsection{Register Load Redundancy Elimination}

Poor memory performance caused by the irregular and redundant memory access is another key bottleneck of efficient DNN execution. \projectname employs two further optimizations to address this challenge: (1) matrix tiling (with the best tiling size decided by auto-tuning) to improve the load/store efficiency from memory to register, and (2) register load redundancy elimination (LRE) to reduce the number of register loads.  
This section focuses on the latter because of its novelty.

\begin{figure}[t]
    \vspace{12pt}
    \centering
    \includegraphics[width=0.47 \textwidth]{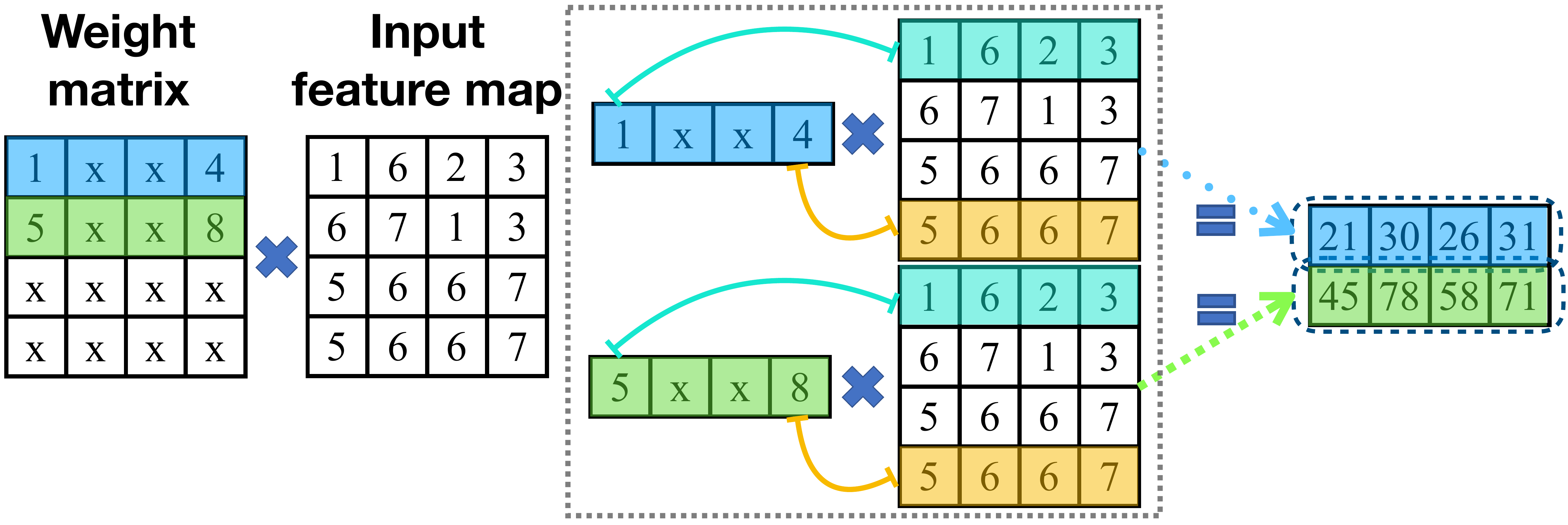}
    \caption{{\bf Register level LRE.}}
    \label{fig:filter_lre}
\end{figure}

Figure~\ref{fig:filter_lre} shows a register-level RLE example, in which both [1,4] and [5,8] (i.e. the first two rows) in the kernel matrix require the first and the last rows of the input feature map. Thus, the first and last rows of the input feature map could be loaded into the register once and reused by the first two rows of the kernel matrix.
\projectname achieves this by a proper loop unrolling transformation (as shown in Figure~\ref{fig:compiler_controlflow}, {\tt +LRE}), because this LRE opportunity is decided by the kernel matrix that is already known during the compilation time. 

It is worth noting that although it is easy to implement this LRE for dense models, it is  challenging (even not possible) for randomly pruned models. Our BCR pruning re-enables LRE, showing the benefit of a model compression and compiler optimization co-design.

\subsection{Auto-Tuning and Other Optimizations}

\projectname also includes some other optimizations that improve execution performance obviously:

\noindent{\bf Auto-tuning.} DNN execution usually involves many configurable performance parameters, such as the data placement on GPU heterogeneous memory, matrix tiling sizes, loop unrolling factors, etc. Tuning them manually is tedious and error-prone. \projectname thus includes an auto-tuning module based on Genetic Algorithm to explore them automatically. In particular, after BCR pruning, different model kernels have varied sizes and shapes that require different tiling shapes and thread block settings. \projectname employs this auto-tuning module to extensively explore the best configurations for all DNN kernels. Comparing to existing auto-tuning approaches in TVM, \projectname's auto-tuning exploits better parallelism because its foundation, Genetic Algorithm allows starting parameter search with initializing an arbitrary number of chromosomes. \projectname's auto-tuning is more efficient.

\noindent{\bf Vectorization.} \projectname also vectorizes CPU and GPU code automatically with ARM NEON and OpenCL, respectively. CPU and GPU have different (and limited) numbers of vector registers. To fully utilize them while minimizing the register spilling, \projectname carefully designs another level of loop unrolling to pack more computations together. Combining this optimization with the regularity given by BCR pruning and matrix reorder, \projectname generates more efficient vector codes comparing to other DNN acceleration frameworks. 

\noindent{\bf Computation Transformation.}
\projectname transforms CONV to sparse matrix multiplication, which  requires to convert CONV weights to a GEMM-based matrix format (i.e, the step of {\tt Im2col} in Figure~\ref{fig:sys_overview}). {\tt Im2col} is memory-bound as it only reads weights and expands them to a larger matrix. \projectname optimizes {\tt Im2col} by skipping the matrix row during expanding, when a certain weight column is completely pruned.

\section{BCR Pruning Optimizations}~\label{sec:BCRpruning}

\vspace{-1em}

In this section, we present the BCR pruning techniques to cooperate with the compiler optimizations.
Besides performing BCR pruning itself, we need to optimize the block size (for each layer) and also other hyperparameters (such as the pruning rate for each layer). 
The search space of all the hyperparameters is huge, therefore we propose a \emph{decoupling strategy} of the hyperparameter space to reduce the problem complexity.
It is based on the following two observations.
First, generally, the sparse DNN accuracy is higher when the block size is smaller.
Second, the mobile inference acceleration relates to the block size (and thus the number of blocks) and is independent of actual weight values. 
Therefore, we decouple block size optimization from other hyperparameter optimizations. 
More specifically, we perform mobile testing with the compiler optimizations to evaluate inference acceleration performance at different block sizes and select the smallest block size such that the inference acceleration performance degradation (compared with pruning whole rows/columns under the same pruning rate) is within a predefined threshold. This step is \emph{independent} of DNN training or actual BCR pruning and should run much faster. The underlying principle is that the derived block size will likely provide the highest accuracy while satisfying the inference acceleration performance requirement. More elaborations about the decoupled optimizations are provided in the following.

\subsection{Block Size Optimization}
\textcolor{revision}{
Block size affects accuracy because a small block size results in a larger search space that mitigates the accuracy loss without changing pruning rate. The granularity of pruning increases with the block size growing. For example, under the same pruning rate, non-structured pruning can achieve higher accuracy than coarse-grained structure pruning. As the block size increases, the BCR pruning with increasing regularity becomes increasingly similar to the coarse-grained structure pruning, while as the block size decreases, the BCR pruning with less regularity becomes increasingly similar to a non-structured pruning. 
}

\lstset{numbers=left, numberstyle=\tiny, stepnumber=1, numbersep=5pt, frame=tb, caption=Block size optimization,emph={},emphstyle=\underbar}
\begin{lstlisting}[label={lst:find-optimal-block},language=python]
def synthesize(origin_layer, rate, size):
  # Copy the layer information, e.g. Weight shape
  layer = copy_layer(origin_layer)
  # Randomly generate weights
  generate_random_weight(layer, rate, size)
  return layer

# <Algorithm Entry>: find optimal block size with a pruning_rate
def find_opt_blk(layer, rate, block_sizes, device):
  opt_size = -1
  opt_latency = INFINITE_MAX
  # Traverse different block size
  for size in block_sizes:
    # Synthesize a new layer
    syn_layer = synthesize(layer, rate, size)
    # Run synthesized layer on mobile device and get latency
    latency = run_layer(syn_layer, device)
    # if improvement of latency is less than a threshold, then stop 
    if opt_latency / latency < threshold:
      break
    # Record a more optimal block size
    opt_latency = latency
    opt_size = size
  # Return the records
  return opt_size
\end{lstlisting}

The block size optimization is based on offline mobile testing with the compiler optimizations. Its goal is to select the smallest block size for each layer such that the inference acceleration performance degradation is within a tolerable range.
\textcolor{revision}{
Different layers may have different properties, e.g. the size of feature maps and convolution kernels. The runtime performance of different layers may vary as different properties result in different cache performance and computation patterns. Therefore, \projectname requires to study the relationship between layer size (and layer structure) and desirable block size and identifies the preferred block size for each layer.
}
We evaluate mobile CPU/GPU in a layerwise manner, \emph{using synthesized BCR pruning strategies with a pruning rate for each target layer}, and then select the desirable block size.

\textcolor{revision}{
List~\ref{lst:find-optimal-block} shows our detailed block size optimization algorithm. The objective of this algorithm is to find the optimal block size for each target layer under certain pruning rates. It consists of three steps. 
The first step accepts a layer, a pruning rate, a candidate block size set, and a device (mobile CPU or GPU) as input for the optimal block size testing (Listing~\ref{lst:find-optimal-block} line 9). The block size set consists of a group of width and height pairs that can divide the layer’s widths and heights. The second step generates a synthesized layer with features (e.g. CONV kernel size, weight size, and stride size, except the real weight numbers) identical to the original layer. This step also generates random weights that satisfy the pruning rate and block size requirements (Listing~\ref{lst:find-optimal-block} lines 1 to 6) to improve the algorithm efficiency by avoiding training weights. The key insight is that the pruning ratio rather than the specific location of non-zero weights impacts more on the latency of a certain CONV layer with our BCR pruning that already guarantees certain regularity. The third step runs this synthesized layer on a mobile device to test its latency. If the latency improvement of this synthesized layer exceeds a threshold, our algorithm sets this block size as local optimal; otherwise it terminates with returning the last local optimal block size (Listing~\ref{lst:find-optimal-block} lines 19 to 25). 
This algorithm executes on a host (PC) machine except the third step ({\em run\_layer}) that runs on mobile devices to test latency. The whole process performs offline efficiently, independent of training/pruning. For example, for VGG-16 (on ImageNet dataset), this process takes less than 1 hours.
}

\begin{figure}[t]
    \centering
        \subfloat[$1024\times 1024$ matrix.]{
            \includegraphics[width=0.225\textwidth]{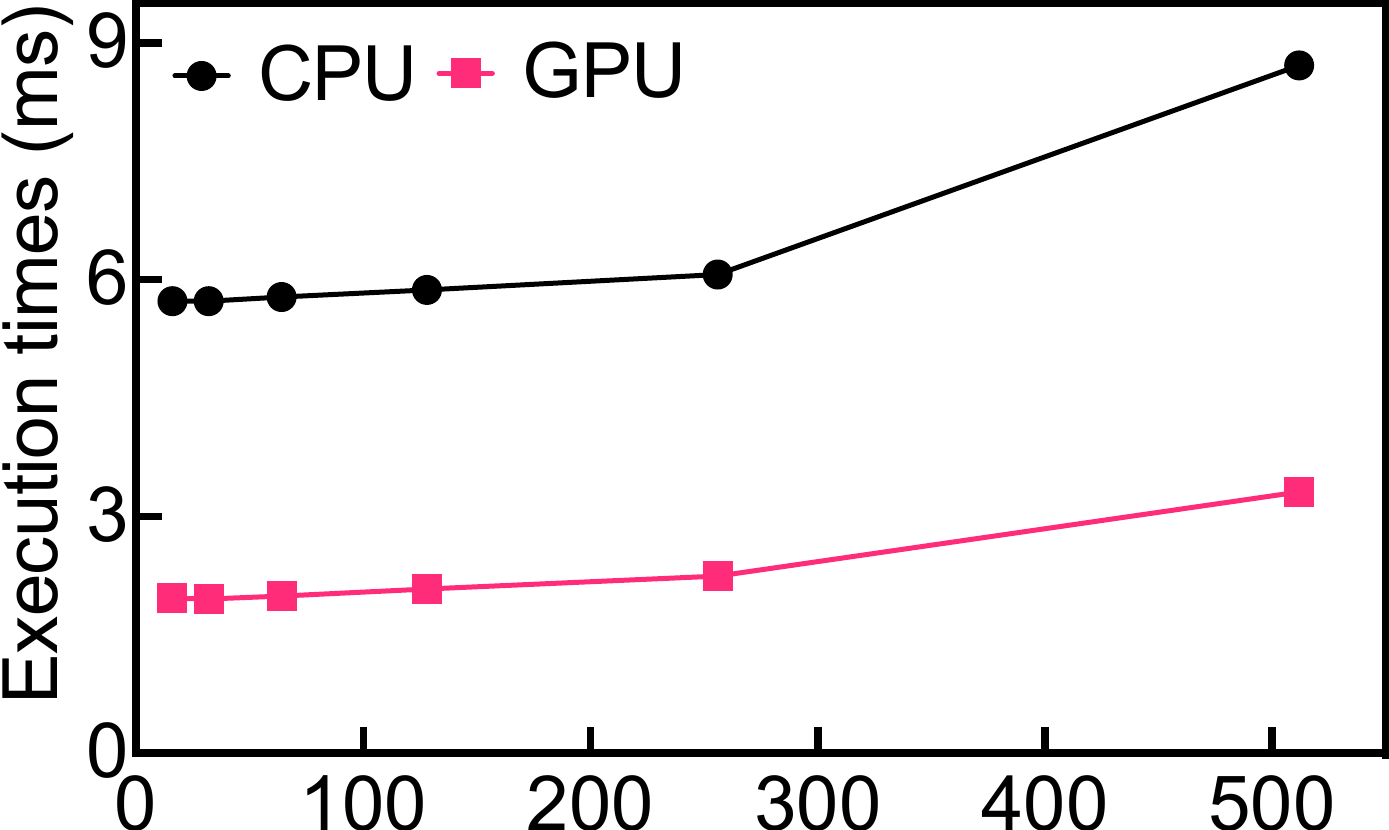}
        }
        \subfloat[Whole VGG-16.]{
            \includegraphics[width=0.233\textwidth]{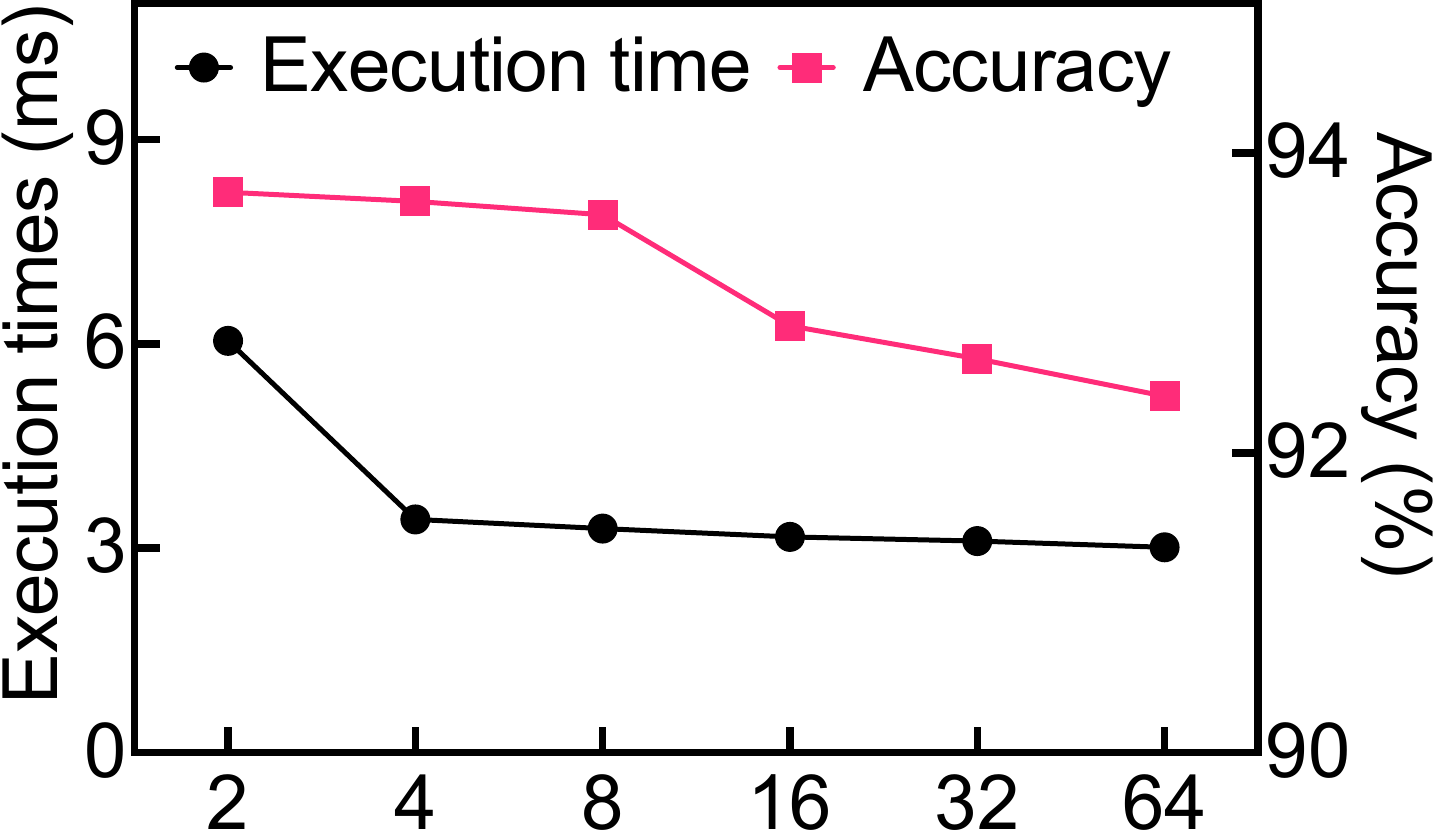}
        }
        \caption{{\bf (a) The CPU and GPU execution time (y-axis) for a single weight matrix as changing the number of blocks (x-axis); (b) The CPU execution time (left y-axis) and accuracy (right y-axis) for VGG-16 on CIFAR-10 as changing the block size.}}
    \label{fig:tradeoff_with_accu}
        \vspace{1em}
\end{figure}

The left part of Figure \ref{fig:tradeoff_with_accu} shows an illustrative example using a $1024\times 1024$ weight matrix, under $10\times$ BCR pruning rate. 
As the block number increases, the execution time remains stable before it reaches 256, and increases dramatically after that. 
The right part of Figure \ref{fig:tradeoff_with_accu} shows the execution time and accuracy trend with changing block size for VGG-16 trained on CIFAR-10. The x-axis shows the first dimension of the block size, and the second dimension is fixed as 16. 
With increasing block size, the execution time drops quickly at first until reaching a relatively stable level (around 3 ms), and the inference accuracy drops slowly at first and quickly after a point. Therefore, it is possible for us to find a specific block size (e.g., $4\times 16$) that yields (near-)optimal execution time without compromising accuracy.

\subsection{BCR Pruning using ADMM}
\label{ADMM}

Based on the derived block size (number) for each layer, we will perform BCR pruning along with the optimizations of the remaining key hyperparameters: target pruning rate for each layer. We adopt state-of-the-art weight pruning algorithm using ADMM (Alternating Direction Methods of Multipliers) and generalize to BCR pruning, for two reasons: (i) It achieves (one of) the highest weight pruning rates satisfying accuracy constraint~\cite{zhang2018systematic,zhang2018adam,ren2019admm}. (ii) The ADMM-based solution framework, when generalized to BCR pruning, can automatically determine the desirable column and row pruning rates for each block given a predefined pruning rate for a whole weight matrix (for a specific layer).

\textbf{BCR Pruning Problem Formulation and ADMM-based Solution:} 
For an $N$-layer DNN of interest, let ${\bf{W}}_{i}$ and ${\bf{b}}_{i}$ denote the weights and biases of the $i$-th layer respectively. 
We minimize the loss function associated with the DNN model, subject to the specific fine-grained structured sparsity constraints (columns and rows in a block are pruned) on the weights in the corresponding layers, i.e.,
\vspace{-0.5em}
\begin{equation}
\label{opt0}
\begin{aligned}
& \underset{ \{{\bf{W}}_{i}\},\{{\bf{b}}_{i} \}}{\text{minimize}}
& & f \big( \{{\bf{W}}_{i}\}_{i=1}^N, \{{\bf{b}}_{i} \}_{i=1}^N \big),
\\ & \text{subject to}
& & {\bf{W}}_{i}\in {\bf{S}}_{i}, \; i = 1, \ldots, N
\end{aligned}
\end{equation}
where ${\bf{S}}_{i}$ is the set of ${\bf{W}}_{i}$ with the sparsity constraint $\alpha_i$.

{\emph{\textbf{Fine-grained structured sparsity by BCR}}}: 
Consider the weight matrix of the $i$-th DNN layer divided into $n \times m$ blocks. The constraint on the weight matrix is that, the ratio of the total number of zero weights in all blocks to the total number of weights is no less than $\alpha_{i}$ (the sparsity constraint). And the zero weights form whole columns and rows.

Corresponding to every set ${\bf{S}}_{i}$, $i = 1, \ldots, N$, we define the indicator function
$
g_{i}({\bf{W}}_{i})=
\begin{cases}
 0 & \text { if } {\bf{W}}_{i}\in {\bf{S}}_{i}, \\ 
 +\infty & \text { otherwise.}
\end{cases}
$. Problem (\ref{opt0}) with constraint cannot be solved directly by classic stochastic gradient descent (SGD) methods~\cite{kingma2014adam} as original DNN training.
However, the ADMM regularization can reforge and separate the problem, then solve them iteratively~\cite{hong2016convergence,liu2018zeroth}. First, we reformulate the problem (\ref{opt0}) as:
\vspace{-1em}
\begin{equation}
\label{admm_form}
\begin{aligned}
& \underset{ \{{\bf{W}}_{i}\},\{{\bf{b}}_{i} \}}{\text{minimize}}
& & f \big( \{{\bf{W}}_{i} \}_{i=1}^N, \{{\bf{b}}_{i} \}_{i=1}^N \big)+\sum_{i=1}^{N} g_{i}({\bf{Z}}_{i}),
\\ & \text{subject to}
& & {\bf{W}}_{i}={\bf{Z}}_{i}, \; i = 1, \ldots, N,
\end{aligned}
\end{equation}
where ${\bf Z}_{i}$ is an auxiliary variable.
Then, with formation of augmented Lagrangian~\cite{boyd2011}, the problem (\ref{admm_form}) can be decomposed into two subproblems \eqref{subproblem_1} and \eqref{subproblem_2},
\vspace{-1em}

\begin{equation}
\label{subproblem_1}
 \underset{ \{{\bf{W}}_{i}\},\{{\bf{b}}_{i} \}}{\text{minimize}}
\ \ \ f \big( \{{\bf{W}}_{i} \}_{i=1}^N, \{{\bf{b}}_{i} \}_{i=1}^N \big)+\sum_{i=1}^{N} \frac{\rho_{i}}{2}  \| {\bf{W}}_{i}-{\bf{Z}}_{i}^{t}+{\bf{U}}_{i}^{t} \|_{F}^{2}, \\
\end{equation}
 \begin{equation}\label{subproblem_2}
 \underset{ \{{\bf{Z}}_{i} \}}{\text{minimize}}
\ \ \ \sum_{i=1}^{N} g_{i}({\bf{Z}}_{i})+\sum_{i=1}^{N} \frac{\rho_{i}}{2} \| {\bf{W}}_{i}^{t+1}-{\bf{Z}}_{i}+{\bf{U}}_{i}^{t} \|_{F}^{2}, \\
\end{equation}
where ${\bf U}_{i}$ denotes dual variable and $t$ is the iteration index,  and we update ${\bf U}_{i}$ in each iteration by ${\bf{U}}_{i}^{t}:={\bf{U}}_{i}^{t-1}+{\bf{W}}_{i}^{t}-{\bf{Z}}_{i}^{t}$. These two will be iteratively solved until convergence.

The first subproblem can be solved by classic SGD. For the second subproblem, the solution is given by
\begin{equation}
{\bf{Z}}_{i}^{t+1} = \prod_{{\bf{S}}_{i}}({\bf{W}}_{i}^{t+1}+{\bf{U}}_{i}^{t}), 
\end{equation}
where $\prod_{{\bf{S}}_{i}}(\star) $ is the Euclidean projection to ${{\bf S}_{i}}$, thereby guarantees weight matrices are subjected to {\emph{the fine-grained structured sparsity by BCR}}.

Whole layer pruning rates are the hyperparameters in the ADMM-based solution framework. We use a straightforward, uniform target pruning rate for all layers in the DNN. This is shown as a valid hyperparameter setting for overall acceleration. More sophisticated hyperparameter determination procedure is possible and is orthogonal to this work.

\section{Evaluation}\label{sec:evaluation}

This section evaluates \projectname by comparing it with TVM~\cite{chen2018tvm}, TFLITE~\cite{TensorFlow-Lite}, MNN~\cite{Ali-MNN}, an optimized sparse matrix implementation (CSR) based on CSR~\cite{greathouse2016clsparse}, and PatDNN~\cite{niu2020patdnn}. 

\subsection{Methodology}

\noindent{\bf Evaluation Objective.} Our evaluation has four objectives: (1) proving BCR pruning results in both high pruning rates and accuracy by comparing it with several state-of-the-art model compression efforts;
(2) demonstrating \projectname runs faster than state-of-the-art end-to-end DNN execution frameworks, achieving real-time execution of mainstream DNNs on mobile devices without accuracy compromise; 
(3) studying the performance impact of \projectname's major compiler optimizations and the underlying reasons for the performance gains; (4) validating \projectname's good portability by comparing it with other frameworks on two other mobile devices.   

\noindent{\bf Models and Datasets.} 
\projectname is evaluated on three mainstream CNNs, VGG-16 (VGG), ResNet-18 (RNT) and Mobile-Net-V2 (MBNT). They are trained and tested on two datasets,  CIFAR-10 and ImageNet. Here, we use 4 $\times$ 16 as the block size.
\textcolor{revision}{
 When pruning and retraining models, the initial learning rate is 1e$-$2 for CIFAR-10, and it is reduced to 1e$-$3 for ImageNet. 
The learning rate is fixed for pruning, while adjusted in retraining with a scheduler following the cosine function.
Besides, the numbers of epochs for pruning of all models on CIFAR-10 and ImageNet are fixed to 400 and 100, respectively, and for retraining are 300 and 100, respectively.
For all models' pruning, the penalty factor ($\rho$) increases exponentially from 1e$-$4 to 1e$-$1.
All training are conducted on NVIDIA Titan RTX GPUs with Ubuntu operating system and the PyTorch~1.3 framework with CUDA~10.1. For training cost, take VGG-16 as an example. 
On CIFAR-10 dataset, pruning (with ADMM optimization) and retraining (with normal DNN training setting) consume 22 seconds/epoch and 20 seconds/epoch on average with one Titan RTX GPU, respectively. On ImageNet dataset, pruning and retraining consume 42 minutes/epoch and 38 minutes/epoch on average with four Titan RTX GPUs, respectively. 
}
\projectname is also evaluated on a popular GRU RNN model that is widely used in previous studies \cite{han2017ese, wang2018c, li2019rnn}. GRU contains 2 GRU layers and about 9.6M parameters. GRU is trained and tested on the TIMIT dataset \cite{garofolo1993timit} that is commonly used for evaluating automatic speech recognition systems.

\noindent{\bf Testbed and Evaluation Setup.}
Our evaluations are conducted on a cell phone, Samsung Galaxy S10 with the latest Qualcomm Snapdragon 855 that consists of a Qualcomm Kryo 485 Octa-core CPU and a Qualcomm Adreno 640 GPU. 
The portability is tested on a Xiaomi POCOPHONE F1 phone with a Qualcomm Snapdragon 845 that consists of a Kryo 385 Octa-core CPU and an Adreno 630 GPU, and an Honor Magic 2 phone with a Kirin 980 that consists of an ARM Octa-core CPU and a Mali-G76 GPU.
All experiments run 50 times on varied inputs with 8 threads on CPU, and all pipelines on GPU. Multiple runs do not vary severely, so we only report the average execution time for readability. 
We tune all runs to their best configurations, e.g., we apply Winograd optimization~\cite{lavin2016fast} for all dense runs, and use 16-bit float point for all GPU runs.

\begin{figure*}[t]
    \centering
        \subfloat[ImageNet-CPU]{
            \includegraphics[width=0.244\textwidth]{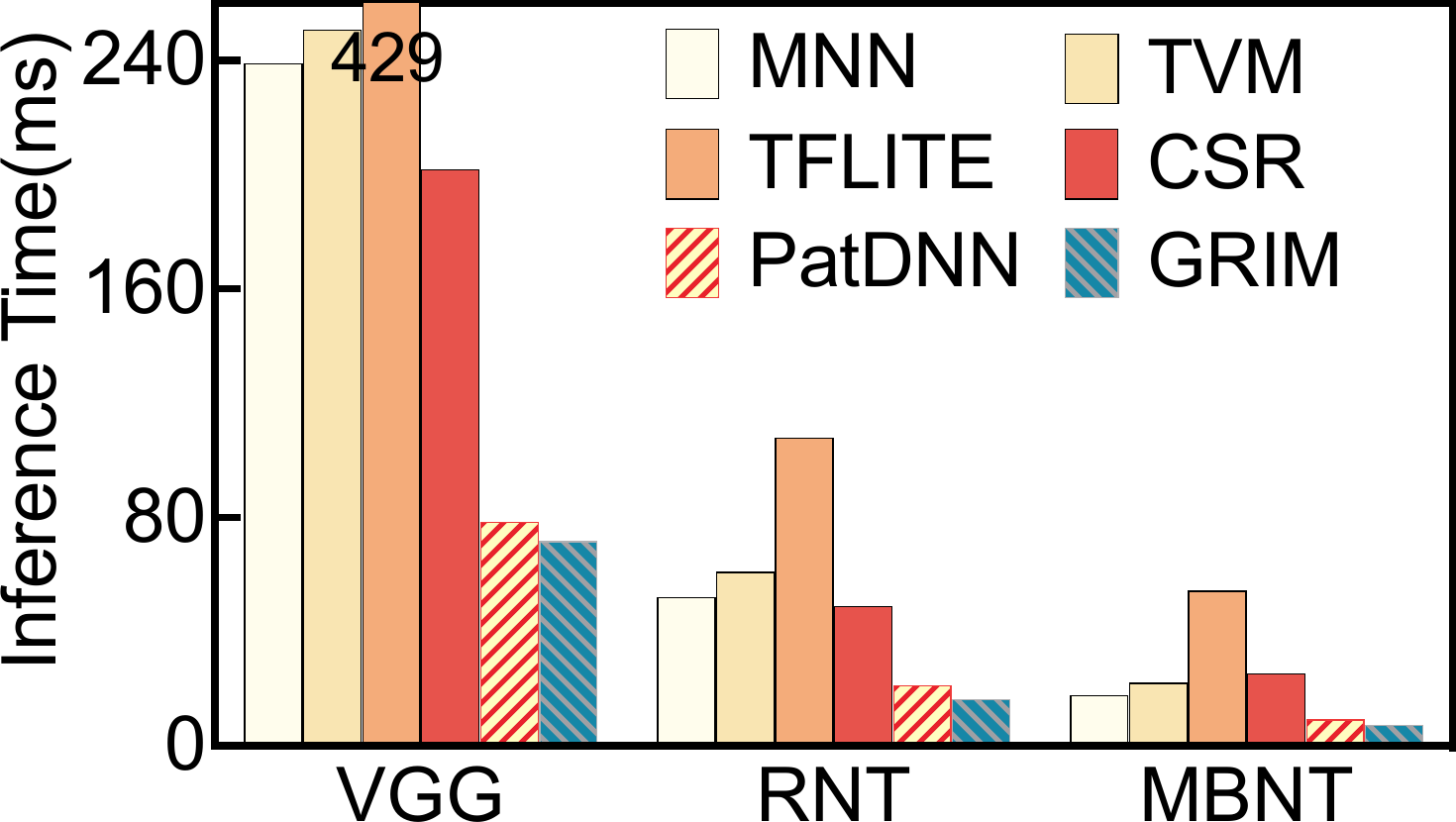}
        }
        \subfloat[ImageNet-GPU]{
            \includegraphics[width=0.24\textwidth]{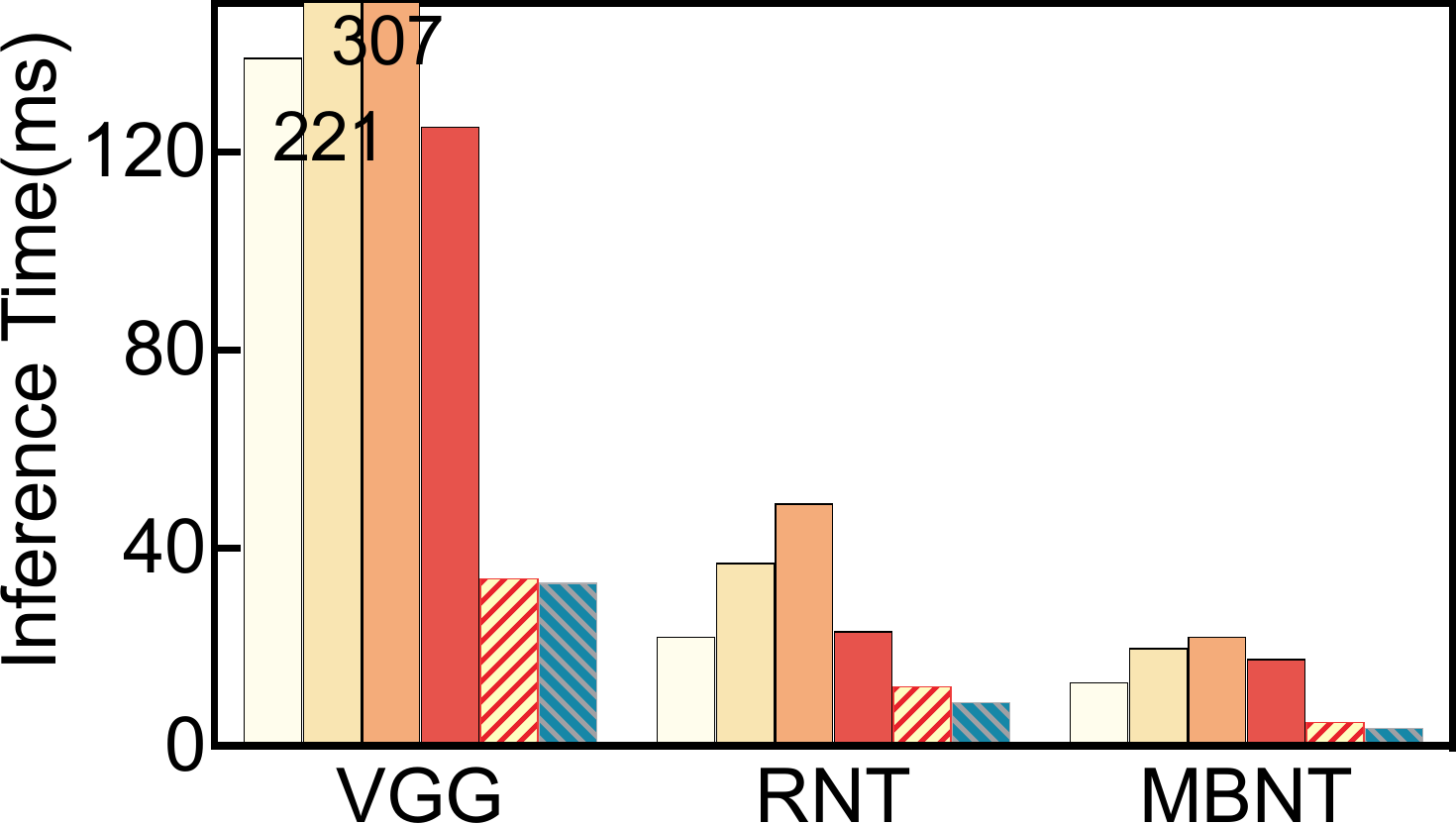}
        }
        \subfloat[CIFAR-10-CPU]{
            \includegraphics[width=0.233\textwidth]{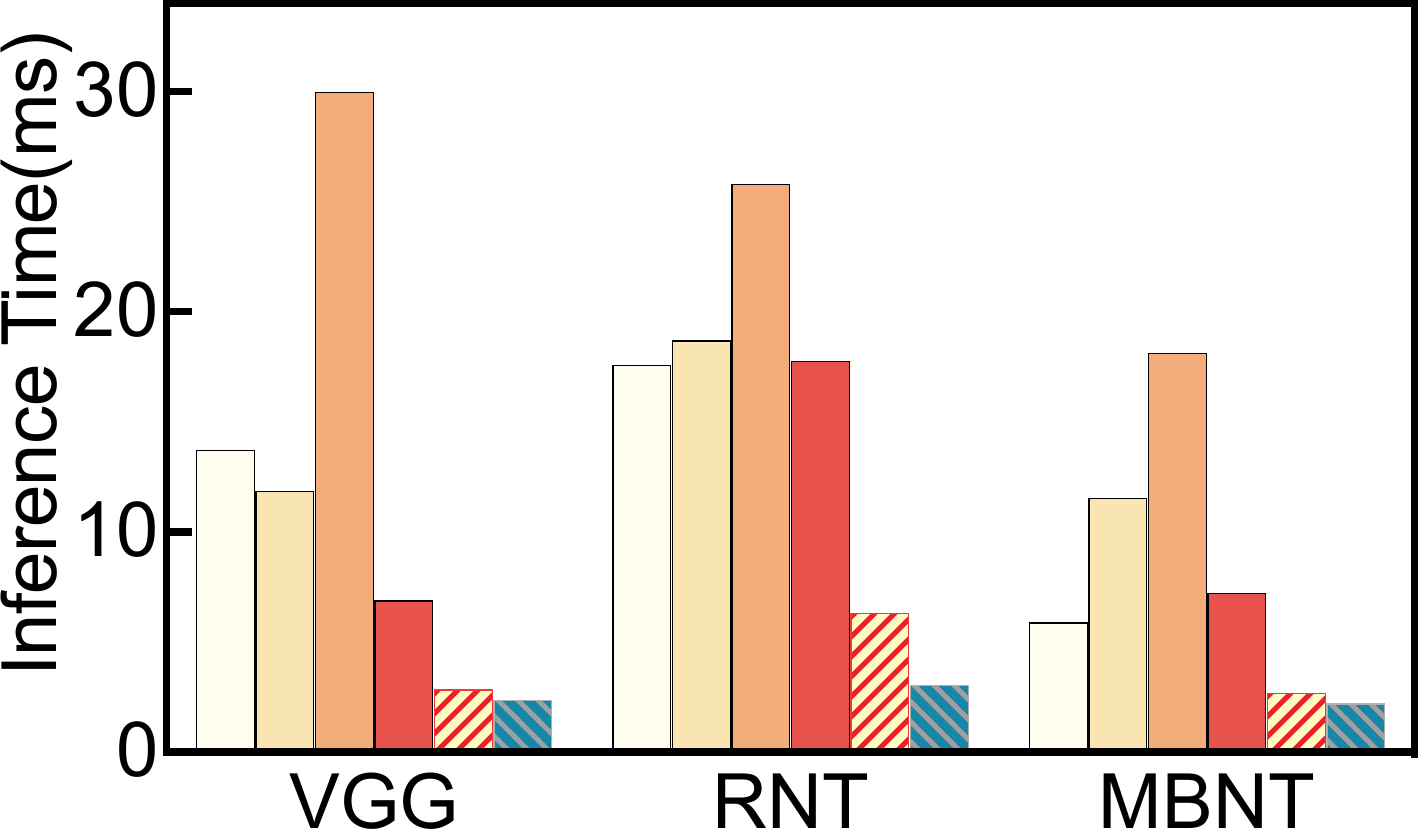}
        }
        \subfloat[CIFAR-10-GPU]{
            \includegraphics[width=0.232\textwidth]{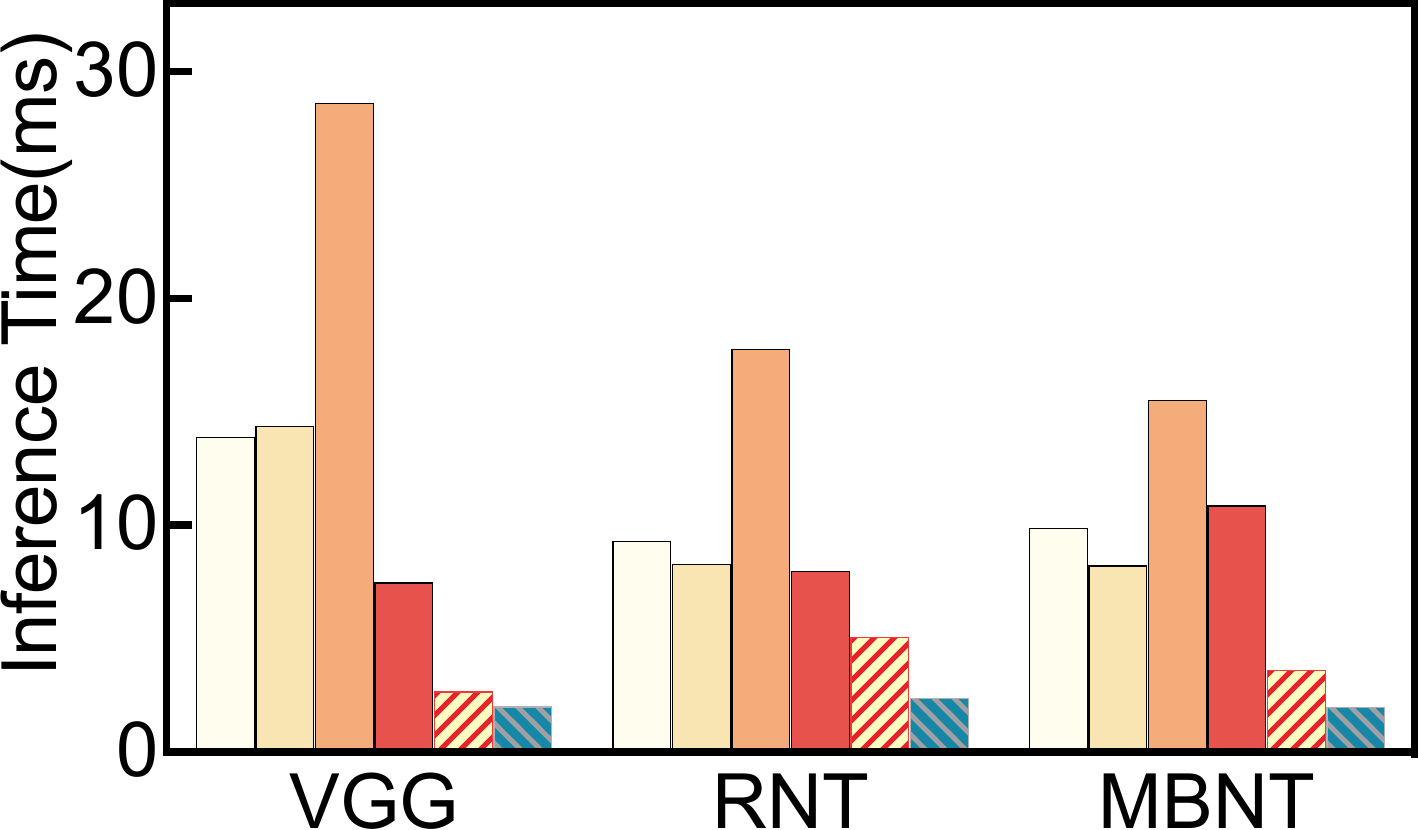}
        }
        \caption{{\bf Overall performance: x-axis are DNN models; y-axis is average DNN end-to-end inference time on a single input.}}
    \label{fig:eva_overview_performance}
        
\end{figure*}

\begin{figure}[t]
\vspace{-1em}
    \centering
        \subfloat[CPU exe time (ms).]{
            \includegraphics[width=0.23\textwidth]{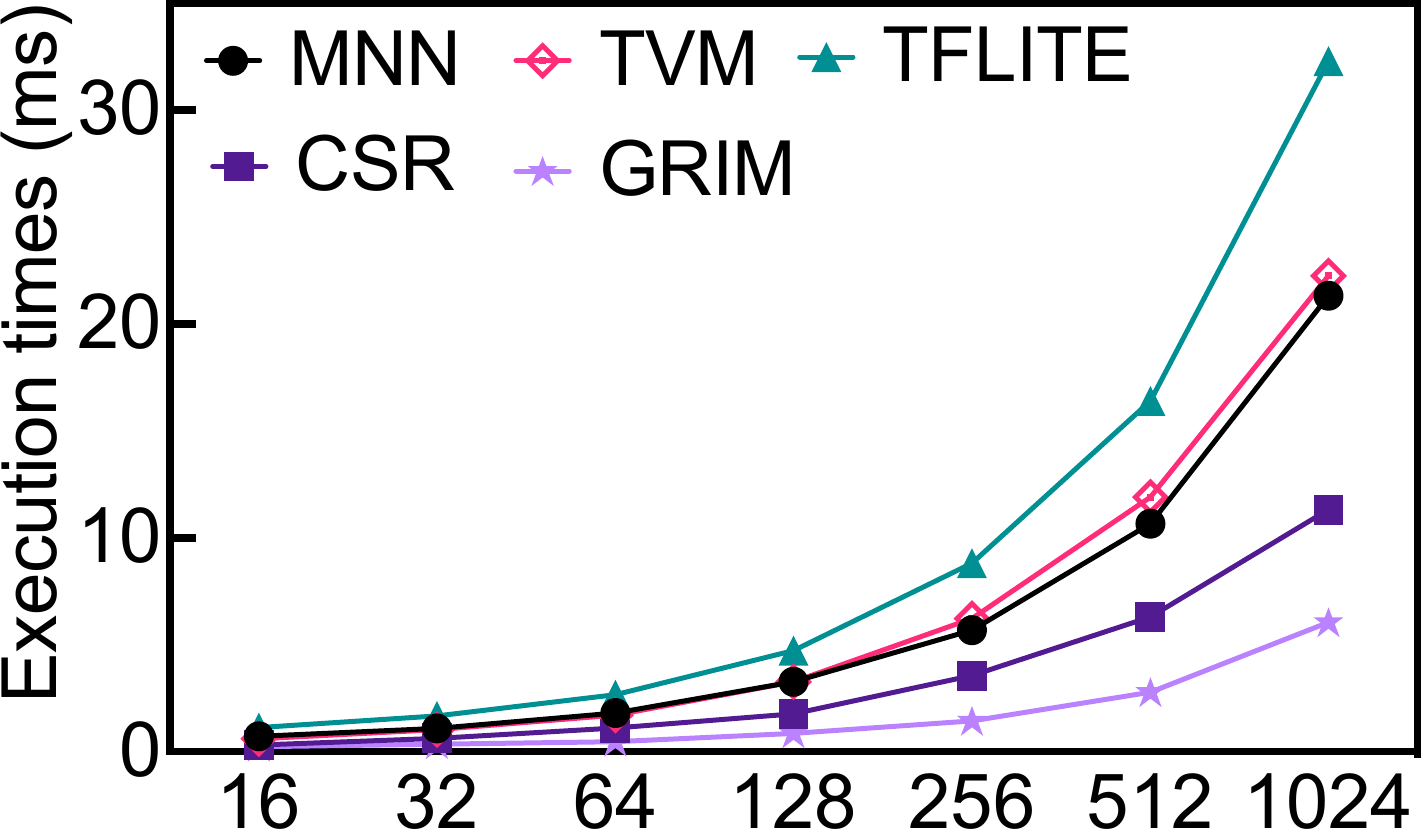}
        }
        \subfloat[GPU exe time (ms).]{
            \includegraphics[width=0.23\textwidth]{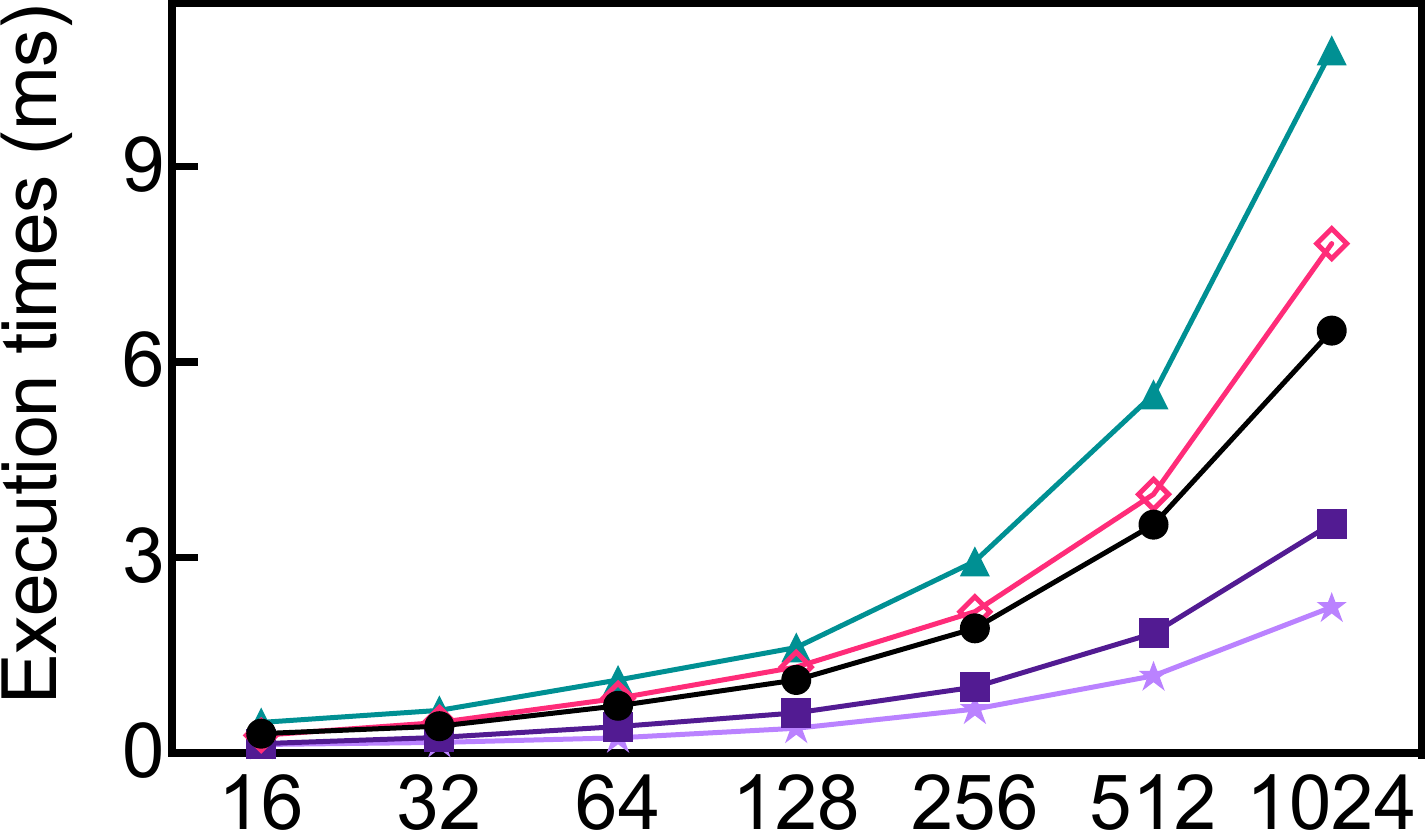}
        }
        \caption{{\bf Matrix Multiply performance: x-axis is row/column size.}}
    \label{fig:eva_gemm}
\end{figure}

\subsection{Accuracy Report}

\emph{\textbf{CIFAR-10.}} Table~\ref{table1} shows the BCR pruning results for CIFAR-10 dataset. We use pre-trained VGG-16, ResNet-18 and MobileNet-V2 networks as our starting points to perform BCR pruning separately. For VGG-16, the original accuracy of the pre-trained model is 93.5\%. Compared to the original model, BCR pruning achieves up to 50.5$\times$ pruning rate without any accuracy degradation. We further extend the pruning rate to 71.3$\times$ and get only 0.4\% accuracy loss. For ResNet-18, when the pruning rate is 24.4$\times$, BCR pruning achieves lossless 94.1\% accuracy. When the pruning rate extends to 27.0$\times$, the accuracy degradation is still negligible. For MobileNet-V2, BCR pruning achieves 9$\times$ pruning rate with minor accuracy loss compared to the original model (94.5\%). Considering MobileNet-V2 is already a compact network, this weight pruning result is still prominent. 
\textcolor{revision}{
To conduct an apple-to-apple comparison among DCP~\cite{zhuang2018discrimination}, PatDNN~\cite{niu2020patdnn}, and \projectname. We also evaluate the accuracy of ResNet-18 under a pruning rate of 27.0$\times$. DCP, PatDNN, and \projectname yield an accuracy of 68.2\%, 92.1\%, and 93.9\%, respectively.
}

\emph{\textbf{ImageNet.}} Table~\ref{table2} shows the BCR pruning results of VGG-16, ResNet-18 and MobileNet-V2 on ImageNet dataset. For VGG, the original model has top-5 accuracy as 91.7\%, and BCR pruning achieves 8$\times$ pruning rate with no accuracy loss for top-5. When the pruning rate reaches 12$\times$, the top-5 accuracy is 90.8\%. For ResNet-18, the accuracy degradation is negligible when the pruning rate is 4$\times$. For MobileNet-V2, BCR pruning achieves 2$\times$ pruning rate with 0.7\% top-5 accuracy degradation.


\begin{table}[!t]\scriptsize
\renewcommand{\arraystretch}{1.2}
\vspace{13pt}
\caption{\textbf{BCR pruning with optimized block size vs. other pruning methods on CIFAR-10.}}
\resizebox{0.48 \textwidth}{!}{
\label{table1}
\begin{tabular}{|ccccc|}
\hline
 Methods & \begin{tabular}[c]{@{}c@{}}Dense \\ Accuracy\end{tabular} & \begin{tabular}[c]{@{}c@{}}Sparse\\ Accuracy\end{tabular} & \begin{tabular}[c]{@{}c@{}}Conv\\ Pruning Rate\end{tabular} & \begin{tabular}[c]{@{}c@{}}Pruning\\ Method\end{tabular} \\ 
 \hline
 \hline
 \multicolumn{5}{|c|}{VGG-16} \\
 \hline
 Iterative Pruning\cite{han2015learning}\cite{liu2018rethinking} & 92.5\% & 92.2\% & 2.0$\times$ & Irregular \\
 One Shot Pruning\cite{liu2018rethinking} & 92.5\% & 92.4\% & 2.5$\times$ & Irregular \\
 2PFPCE\cite{min20182pfpce} & 92.9\% & 92.8\% & 4.0$\times$ & Filter \\
 Efficient ConvNet \cite{li2017pruning} & 93.2\% & 93.4\% & 2.7$\times$ & Filter \\
 \textcolor{revision}{2:4 Pattern \cite{NVIDIA2-4}} & 93.5\% & 93.8\% & 35.7$\times$ & Irregular \\
 \textcolor{revision}{2:4 Pattern \cite{NVIDIA2-4}} & 93.5\% & 93.3\% & 71.3$\times$ & Irregular \\
 PatDNN~\cite{niu2020patdnn} & 93.5\% & 93.7\% & 19.8$\times$ & Pattern-based \\
 \textbf{GRIM} & \textbf{93.5\%} & \textbf{93.8\%} & \textbf{35.7$\times$} & \textbf{BCR} \\ 
 \textbf{GRIM} & \textbf{93.5\%} & \textbf{93.6\%} & \textbf{50.5$\times$} & \textbf{BCR} \\ 
 \textbf{GRIM} & \textbf{93.5\%} & \textbf{93.1\%} & \textbf{71.3$\times$} & \textbf{BCR} \\ 
 \hline
 \hline
 \multicolumn{5}{|c|}{ResNet-18} \\
 \hline
 DCP\cite{zhuang2018discrimination} & 88.9\% & 87.6\% & 2.0$\times$ & Filter \\
 AMC\cite{he2018amc} & 90.5\% & 90.2\% & 2.0$\times$ & Filter \\
 Variational Pruning\cite{zhao2019variational} & 92.0\% & 91.7\% & 1.6$\times$ & Filter \\
PatDNN~\cite{niu2020patdnn} & 94.1\% & 94.2\% & 16.0$\times$ & Pattern-based \\
 \textbf{GRIM} & \textbf{94.1\%} & \textbf{94.4\%} & \textbf{22.9$\times$} & \textbf{BCR} \\
 \textbf{GRIM} & \textbf{94.1\%} & \textbf{94.1\%} & \textbf{24.4$\times$} & \textbf{BCR} \\
 \textbf{GRIM} & \textbf{94.1\%} & \textbf{93.9\%} & \textbf{27.0$\times$} & \textbf{BCR} \\ 
 \hline
 \hline
 \multicolumn{5}{|c|}{MobileNet-V2} \\
 \hline
 DCP\cite{zhuang2018discrimination} & 94.5\% & 94.7\% & 1.4$\times$ & Filter \\
 \textbf{GRIM} & \textbf{94.5\%} & \textbf{94.7\%} & \textbf{6.0$\times$} & \textbf{BCR} \\
 \textbf{GRIM} & \textbf{94.5\%} & \textbf{94.5\%} & \textbf{7.2$\times$} & \textbf{BCR} \\
 \textbf{GRIM} & \textbf{94.5\%} & \textbf{94.4\%} & \textbf{9.0$\times$} & \textbf{BCR} \\ 
 \textbf{GRIM} & \textbf{94.5\%} & \textbf{93.3\%} & \textbf{11.9$\times$} & \textbf{BCR} \\ 
 \hline
\end{tabular}
}
\vspace{-3.0ex}
\end{table}
\begin{table}[!t]
\renewcommand{\arraystretch}{1.2}
\vspace{13pt}
\caption{\textbf{BCR pruning with optimized block size vs. other pruning methods on ImageNet. }}
\resizebox{0.48 \textwidth}{!}{
\label{table2}
\begin{tabular}{|ccccc|}
\hline
 Methods & \begin{tabular}[c]{@{}c@{}}Dense \\ Top 1/5 \\ Accuracy\end{tabular} & \begin{tabular}[c]{@{}c@{}}Sparse\\ Top 1/5 \\ Accuracy\end{tabular} & \begin{tabular}[c]{@{}c@{}}Conv\\ Pruning Rate\end{tabular} & \begin{tabular}[c]{@{}c@{}}Pruning\\ Method\end{tabular} \\ 
 \hline
 \hline
 \multicolumn{5}{|c|}{VGG-16} \\
 \hline
 Decorrelation \cite{zhu2018improving} & 73.1\%/\emph{N/A} & 73.2\%/\emph{N/A} & 3.9$\times$ & Filter \\
 APoZ~\cite{hu2016network} & \emph{N/A}/88.4\% & 66.2/87.6\% & 2.0$\times$ & Filter \\
 \textcolor{revision}{2:4 Pattern \cite{NVIDIA2-4}} & 74.5\%/91.7\% & 73.4\%/91.0\% & 12.0$\times$ & Irregular \\
PatDNN~\cite{niu2020patdnn} & \emph{N/A}/91.7\% & \emph{N/A}/91.6\% & 8.0$\times$ & Pattern-based \\
 \textbf{GRIM} & \textbf{74.5\%/91.7\%} & \textbf{74.4\%/91.7\%} & \textbf{3.0$\times$} & \textbf{BCR} \\
 \textbf{GRIM} & \textbf{74.5\%/91.7\%} & \textbf{74.1\%/91.7\%} & \textbf{8.0$\times$} & \textbf{BCR} \\ 
 \textbf{GRIM} & \textbf{74.5\%/91.7\%} & \textbf{73.1\%/90.8\%} & \textbf{12.0$\times$} & \textbf{BCR} \\ 
 \hline
 \hline
 \multicolumn{5}{|c|}{ResNet-18} \\
 \hline
 Network Slimming~\cite{liu2017learning} & 68.9/88.7\% & 67.2/87.4\% & 1.4$\times$ & Filter \\
DCP~\cite{zhuang2018discrimination} & 69.6/88.9\% & 64.1/85.7\% & 3.3$\times$ & Filter \\
PatDNN~\cite{niu2020patdnn} & 69.9/89.1\% & 69.5/89.2\% & 4.0$\times$ & Pattern-based \\
 \textbf{GRIM} & \textbf{69.9\%/89.1\%} & \textbf{69.6\%/89.2\%} & \textbf{4.0$\times$} & \textbf{BCR} \\
 \textbf{GRIM} & \textbf{69.9\%/89.1\%} & \textbf{68.4\%/88.6\%} & \textbf{6.0$\times$} & \textbf{BCR} \\
 \textbf{GRIM} & \textbf{69.9\%/89.1\%} & \textbf{67.2\%/87.7\%} & \textbf{8.0$\times$} & \textbf{BCR} \\ 
 \hline
 \hline
 \multicolumn{5}{|c|}{MobileNet-V2} \\
 \hline
 AMC~\cite{he2018amc} & 71.8\%/\emph{N/A} & 70.8\%/\emph{N/A} & 1.4$\times$ & Irregular \\
 \textbf{GRIM} & \textbf{70.9\%/90.4\%} & \textbf{70.0\%/89.7\%} & \textbf{2.0$\times$} & \textbf{BCR} \\
 \hline
\end{tabular}
}
\vspace{+1.0ex}
\end{table}

\begin{table}[!t]
\renewcommand{\arraystretch}{1.2}
\caption{\textbf{BCR pruning with optimized block size vs. other methods on TIMIT. PER is {\em phone error rate.}}}
\resizebox{0.48 \textwidth}{!}{
\label{table3}
\begin{tabular}{|ccccc|}
\hline
 Methods & \begin{tabular}[c]{@{}c@{}}Dense \\ PER\end{tabular} & \begin{tabular}[c]{@{}c@{}}Sparse\\ PER\end{tabular} & \begin{tabular}[c]{@{}c@{}}Pruning Rate\end{tabular} & \begin{tabular}[c]{@{}c@{}}Pruning\\ Method\end{tabular} \\ 
 \hline
 \hline
 \multicolumn{5}{|c|}{GRU} \\
 \hline
 ESE\cite{han2017ese} & 20.40\% & 20.70\% & 8.0$\times$ & Irregular \\
 C-LSTM\cite{wang2018c} & 24.15\% & 24.57\% & 8.0$\times$ & Block-circulant \\
 C-LSTM\cite{wang2018c} & 24.15\% &  25.48\% & 16.0$\times$ & Block-circulant \\
 E-RNN  \cite{li2019rnn} & 20.02\% & 20.20\% & 8.0$\times$ & Block-circulant \\
 \textbf{GRIM} & \textbf{18.8\%} & \textbf{18.8\%} & \textbf{10.0$\times$} & \textbf{BCR} \\ 
 \textbf{GRIM} & \textbf{18.8\%} & \textbf{18.8\%} & \textbf{19.5$\times$} & \textbf{BCR} \\ 
 \textbf{GRIM} & \textbf{18.8\%} & \textbf{23.2\%} & \textbf{103.8$\times$} & \textbf{BCR} \\ 
 \textbf{GRIM} & \textbf{18.8\%} & \textbf{24.2\%} & \textbf{245.5$\times$} & \textbf{BCR} \\
 \hline
\end{tabular}
}\vspace{-2.0ex}
\end{table}

\emph{\textbf{TIMIT for RNN.}} Table~\ref{table3} shows the BCR pruning results that are evaluated by phone error rate (PER) and pruning rate. We compare BCR pruning with other state-of-the-art methods, including ESE\cite{han2017ese}, C-LSTM\cite{wang2018c} and E-RNN\cite{li2019rnn} on the same dataset TIMIT. When pruning rates are low (i.e., not higher than 20$\times$), the BCR pruning guarantees no accuracy degradation, which outperforms ESE at 8$\times$ pruning rate and C-LSTM at both 8$\times$ and 16$\times$ pruning rates in terms of both pruning rate and accuracy. When pruning rates are high (such as 103.8$\times$), BCR can maintain an admirable speech recognition performance, which means the BCR pruned model can even outperform C-LSTM regarding both pruning rate and accuracy. Moreover, the BCR method can well adapt to \textit{ultra-high pruning rate} scenario, e.g., our model with 245$\times$ pruning rate can still maintain a comparable phone error rate (24.20\%) to C-LSTM (24.15\%).


\subsection{Overall Execution Time Report}

Figure~\ref{fig:eva_overview_performance} reports \projectname's CPU and GPU execution performance, and compares \projectname with MNN \cite{Ali-MNN}, TVM \cite{chen2018tvm}, TFLITE \cite{TensorFlow-Lite}, CSR \cite{greathouse2016clsparse}, and PatDNN \cite{niu2020patdnn} on three CNNs (VGG-16, ResNet-18, and MobileNet-V2) trained on two datasets (ImageNet and CIFAR-10), respectively\footnote{Sparse models w/ highest pruning rate in Table ~\ref{table1} to ~\ref{table3} are selected}. 
These evaluations test and report the whole model execution time rather than the time of CONV layers only as PatDNN.
\projectname outperforms other frameworks for all cases. On CPU, \projectname achieves 2.47$\times$ to 5.75$\times$, 3.08$\times$ to 6.11$\times$, 5.98$\times$ to 12.55$\times$, 2.81$\times$ to 5.81$\times$, and $1.09\times$ to $2.06\times$ speedup over MNN, TVM, TFLITE, CSR, and PatDNN, respectively. On GPU, \projectname achieves 2.46$\times$ to 6.84$\times$, 3.08 to 7.09$\times$, 5.47$\times$ to 14.08$\times$, 2.59$\times$ to 5.41$\times$, and $1.03\times$ to $2.11\times$ speedup over MNN, TVM, TFLITE, CSR, and PatDNN, respectively. 
Particularly, comparing to PatDNN, GRIM's BCR pruning is more flexible, not only working for $3\times3$ (or $5\times5$, etc.) {\tt CONV} layers targeted by PatDNN but also leading to better pruning and proper optimizations for $1\times1$ {\tt CONV} and varied {\tt FC} layers that PatDNN cannot fully optimize.
For the largest CNN (VGG) trained on the largest dataset (ImageNet), \projectname can complete the whole inference of a single input within 33 ms with our mobile GPU, meeting the industrial real-time standard (i.e., 30 frames/sec). 

\textcolor{revision}{
BCR pruning requires to convert all convolutions to GEMM through {\tt Im2col}, which incurs overhead, particularly for large kernels.
To validate \projectname's performance on large kernels, this part compares the performance of two kernel sizes, (3, 3) and (11, 11) under the same computation workload (by changing the number of channels) and a 10$\times$ pruning rate. The (3, 3) CONV layer achieves $4.5\times$ speedup over TFLite, while (11, 11) achieves $3.3\times$ speedup over TFLite. 
This is consistent with our intuition and proves that \projectname can still result in notable performance gains on larger kernels regardless of the overhead introduced by {\tt Im2col}. Moreover, large kernels ($>$ 5x5) usually occupy a small portion of the overall computation of models (e.g. 9.3\% in AlexNet). 
}

\textcolor{revision}{
To further validate the advantages of \projectname, this part also compares it with another cutting-edge sparsification approach (named 2:4 pattern) proposed by NVIDIA\cite{NVIDIA2-4}. 2:4 pattern can also be defined as a fine-grained structured pruning that is natively supported by the latest NVIDIA GPU architecture. This work implements 2:4 pattern pruning with the algorithm stated in the section~\ref{ADMM}. Table ~\ref{table1} and Table ~\ref{table2} shows that \projectname achieves comparable accuracy with 2:4 pattern pruning under the same pruning rate. 
Because neither mobile CPU or mobile GPU is equipped with dedicated hardware that supports 2:4 pattern pruning, this work uses the CSR baseline aforementioned to store the sparse data and performs the computation of 2:4 pattern. For VGG-16 on CIFAR-10 (with $71.3\times$ pruning rate) and VGG-16 on ImageNet (with $12\times$ pruning rate), \projectname achieves $2.6\times$ and $3.1\times$ speedup compared with this CSR-based NVIDIA 2:4 implementation on mobile CPU, respectively.
}

%

For GRU RNN, because the above mobile frameworks do not support end-to-end execution. We compare \projectname with them on matrix multiplication kernels with varied sizes.

The weight matrix is pruned with a $10\times$ pruning rate.  Figure~\ref{fig:eva_gemm} reports the result. All frameworks' execution time increases as the matrix size grows. \projectname performs the best, with up to 2.3$\times$, 4.3$\times$, 6.1$\times$, and 2.5$\times$ speedup over MNN, TVM, TFLITE, and CSR. PatDNN is not listed because it optimizes CONV directly without transforming to GEMM. \projectname completes GRU inference on Adreno 640 GPU within 81$us$ (for sequence length of 1 and batch size of 32). We compare \projectname with a representative FPGA implementation, ESE\cite{han2015learning}. \projectname can even slightly outperform ESE\footnote{ESE completes GRU with around 82$us$}.
Specifically, \projectname can achieve significantly higher energy efficiency ($38\times)$ comparing with ESE.

\begin{figure}[t]
    \centering
        \subfloat[ImageNet-VGG16-CPU]{
            \includegraphics[width=0.225\textwidth]{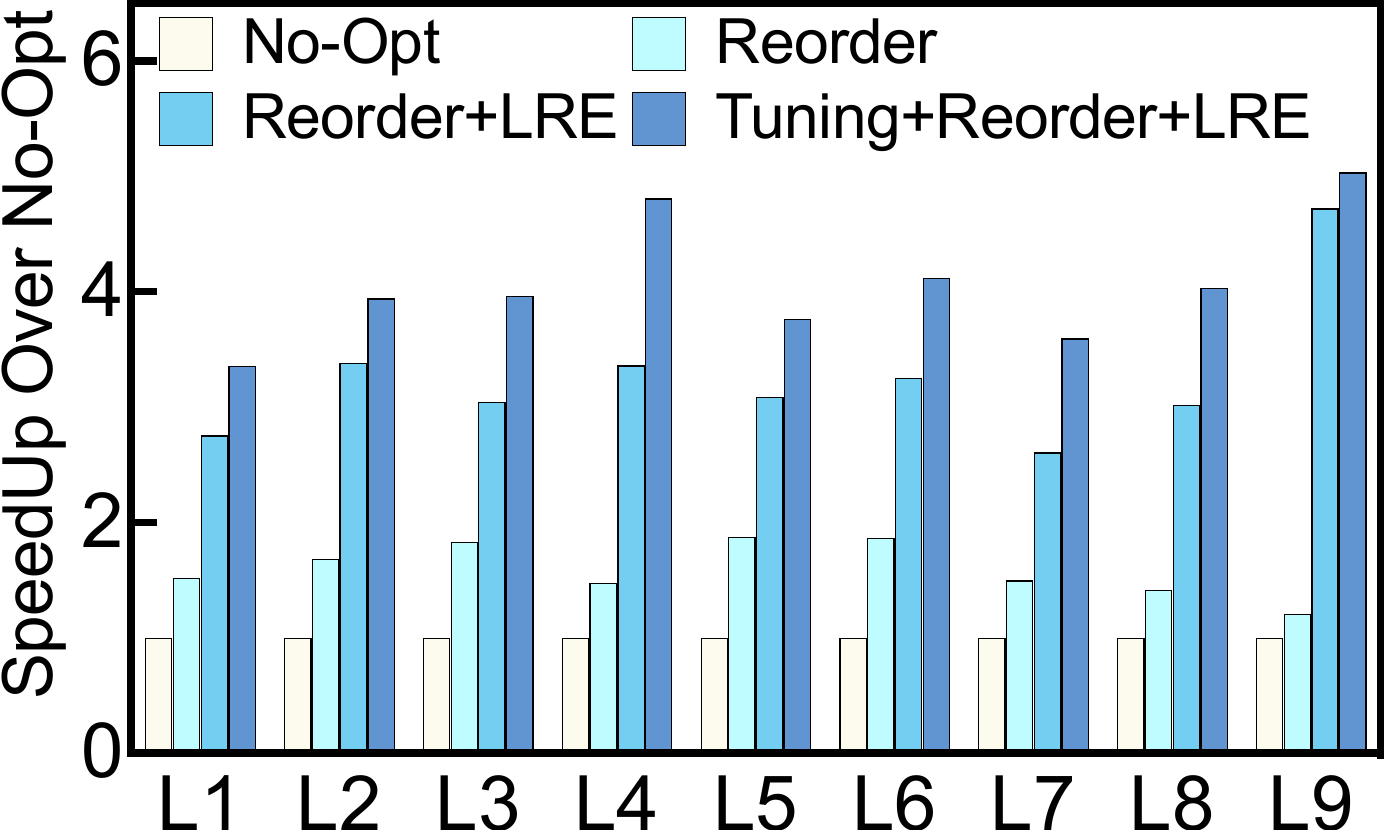}
        }
        \subfloat[ImageNet-VGG16-GPU]{
            \includegraphics[width=0.225\textwidth]{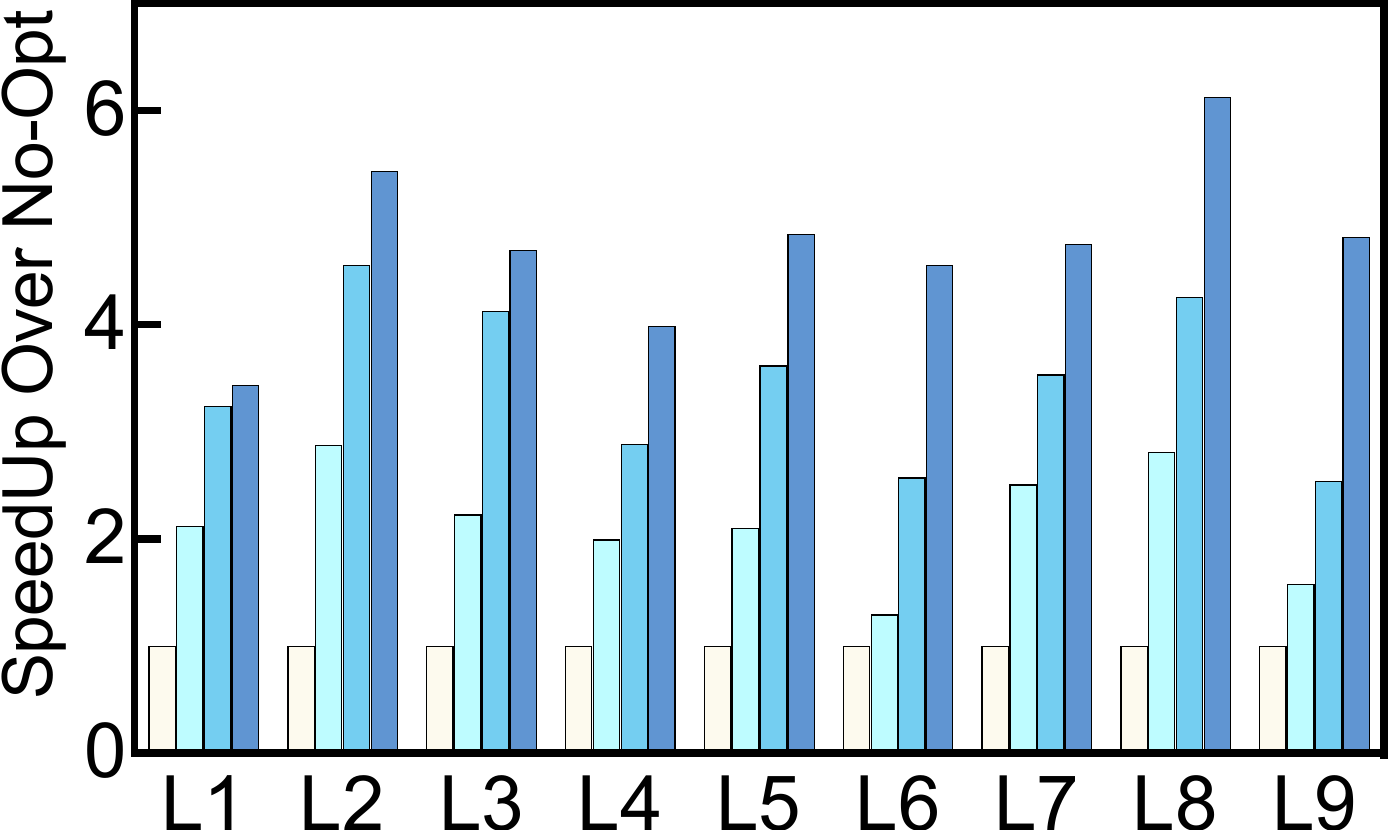}
        }
        \caption{{\bf Speedup: Opt  over No-Opt on different CONV layers of VGG.}}
    \label{fig:eva_optimization}
\end{figure}

\begin{figure}[t]
    \vspace{-8pt}
    \centering
        \subfloat[RNN: a 1024 $\times$ 1024 layer (GRU).]{ 
            \includegraphics[width=0.225\textwidth]{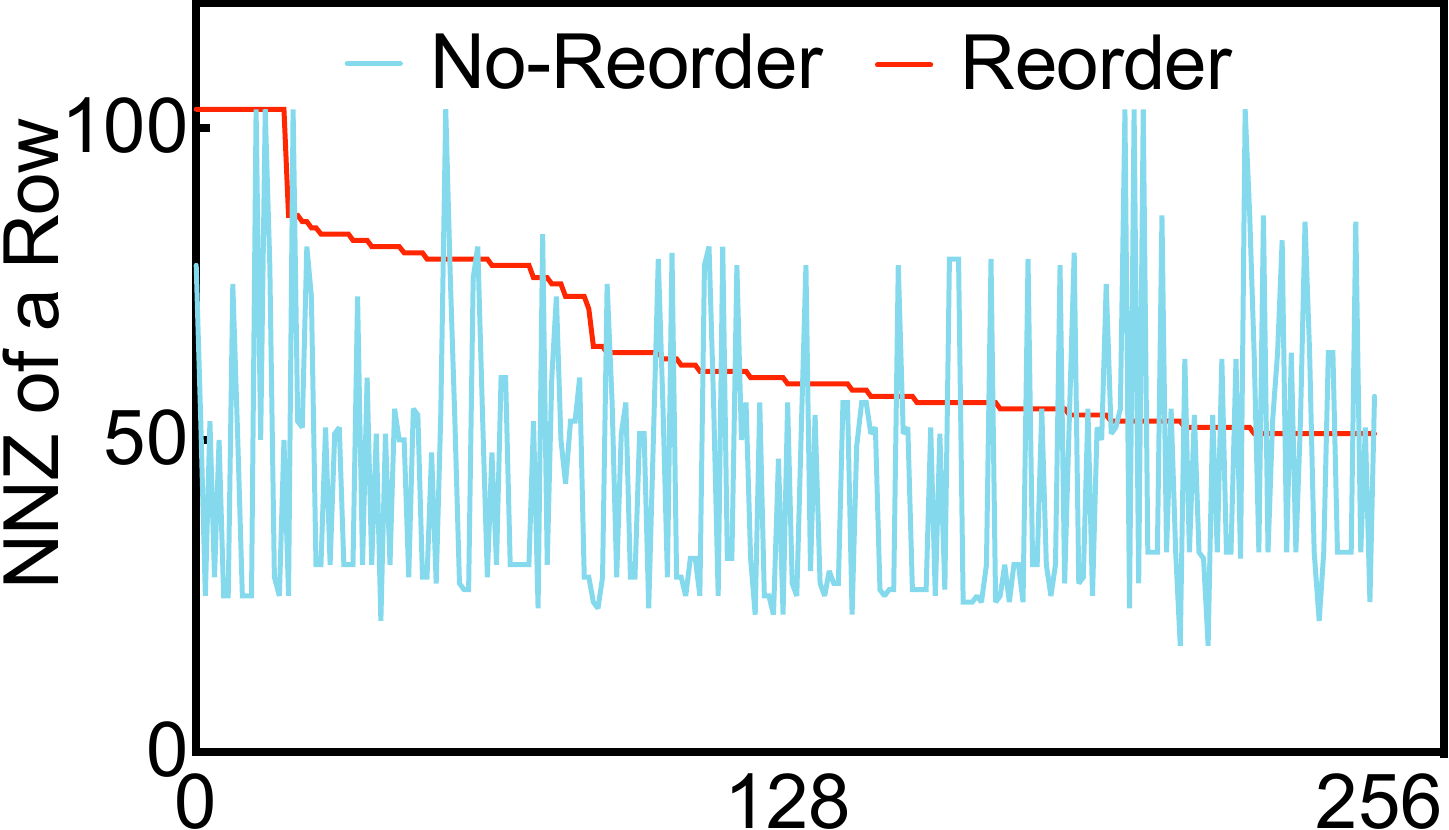}
        }
        \subfloat[CNN: {256$\times$128} in/out channels (VGG).]{ 
            \includegraphics[width=0.225\textwidth]{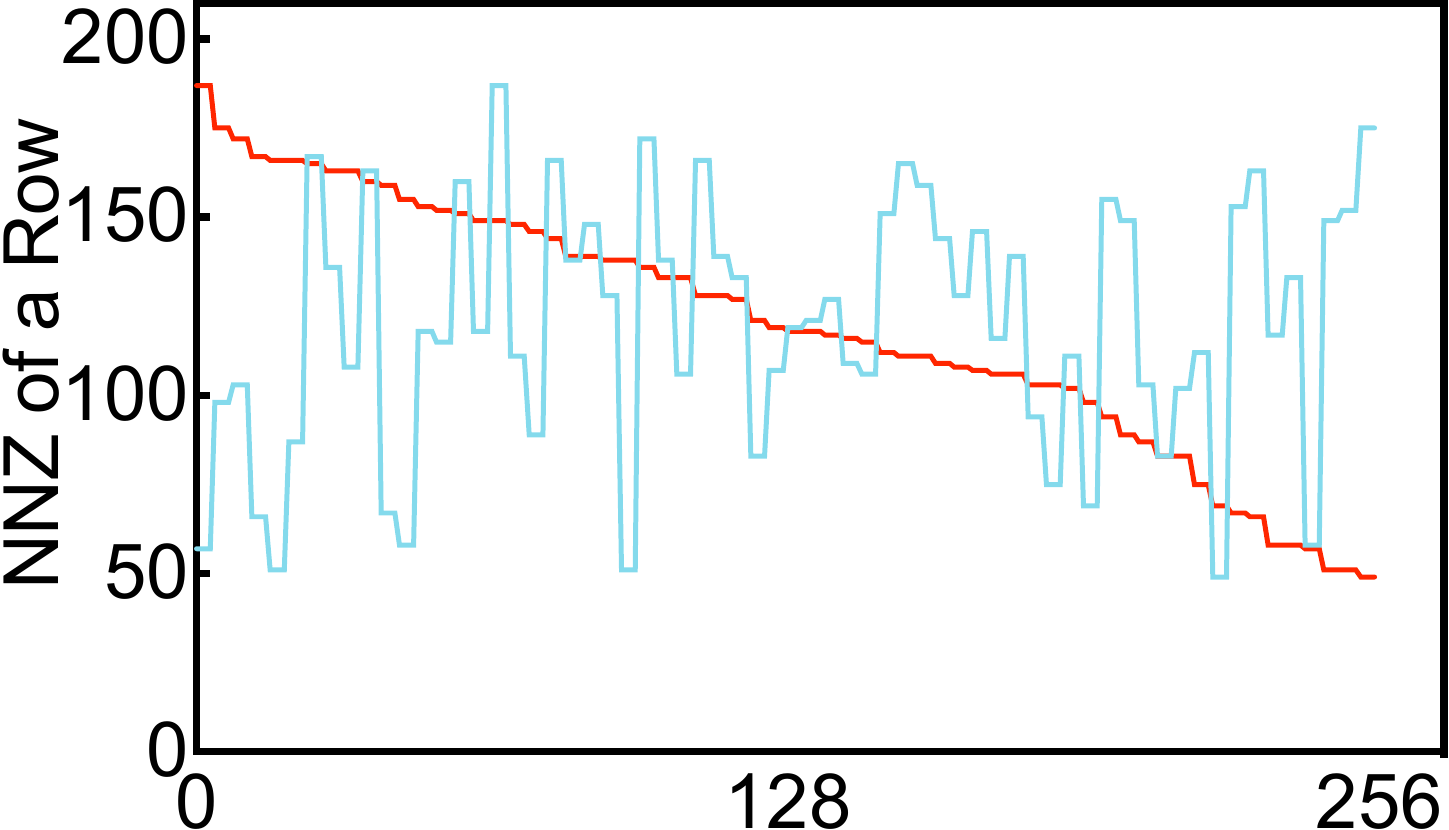}
        }
        \caption{{\bf Matrix reorder: x-axis is row id.}}
    \label{fig:eva_matrix_reorder}
        \vspace{-0.5em}
\end{figure}    

\begin{figure}[t]
    \centering
        \subfloat[GRU(RNN)]{
            \includegraphics[width=0.23\textwidth]{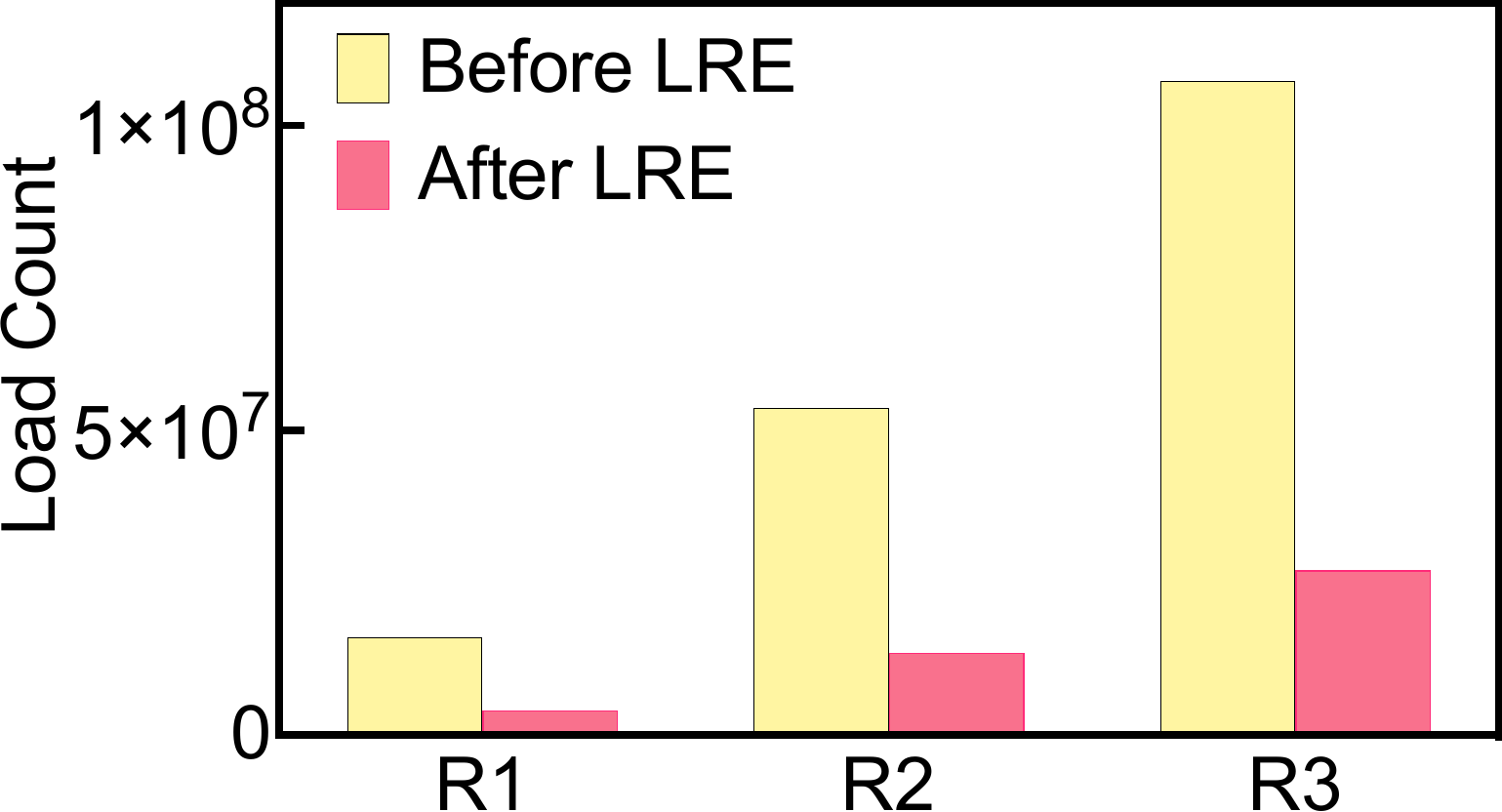}
        }
        \subfloat[VGG(CNN)]{
            \includegraphics[width=0.23\textwidth]{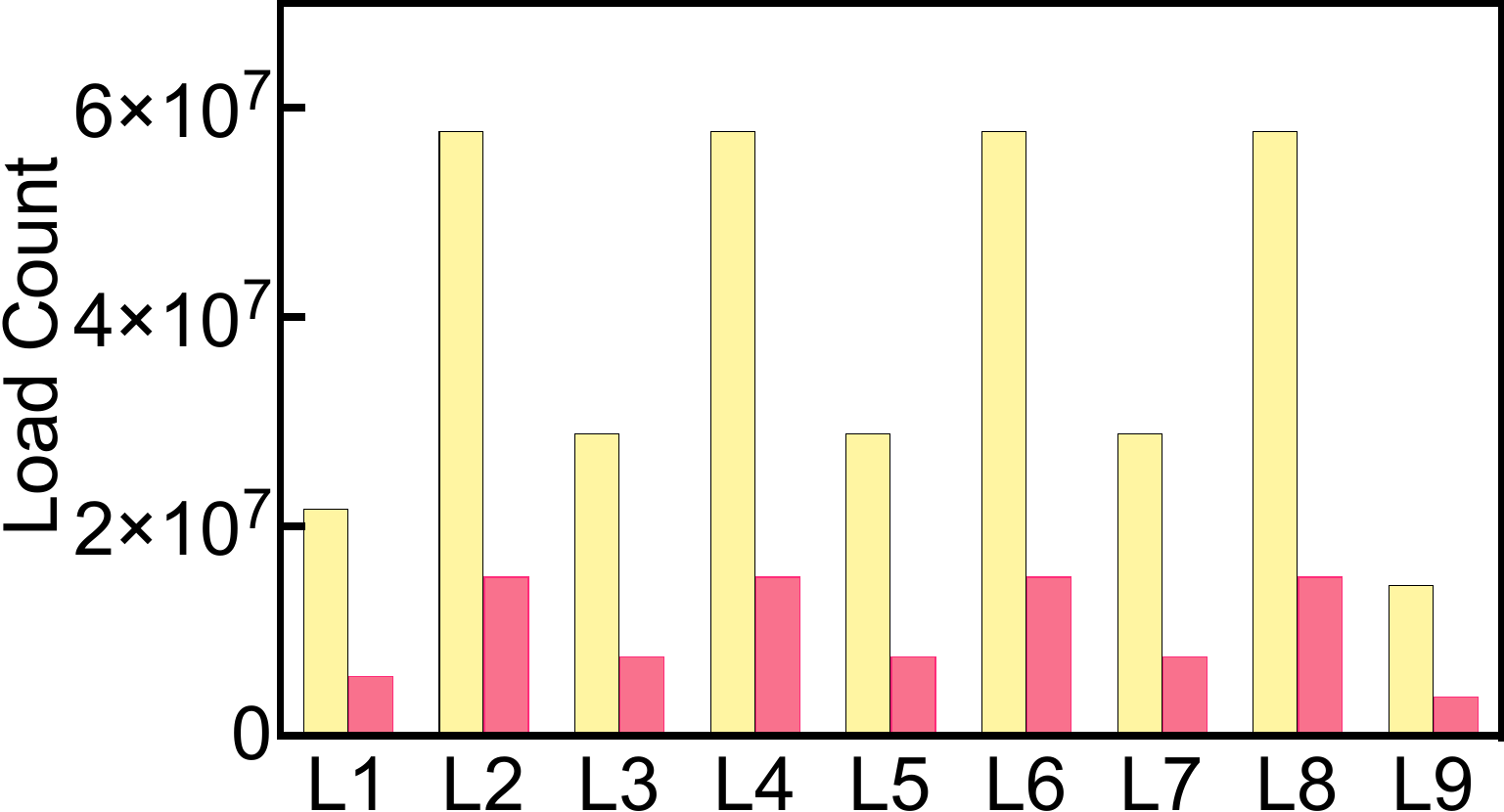}
        }
        \caption{{\bf Register load counts before and after LRE. (R1 to R3 in RNN are layers of GRU with different matrix sizes, $152\times 1024$, $512\times 1024$, $1024\times 1024$. CNN uses CONV layers from VGG.)}}
    \label{fig:eva_lre_count}
    \vspace{0.5em}
\end{figure}

\begin{figure}[t]
    \centering
    \includegraphics[width=0.47 \textwidth]{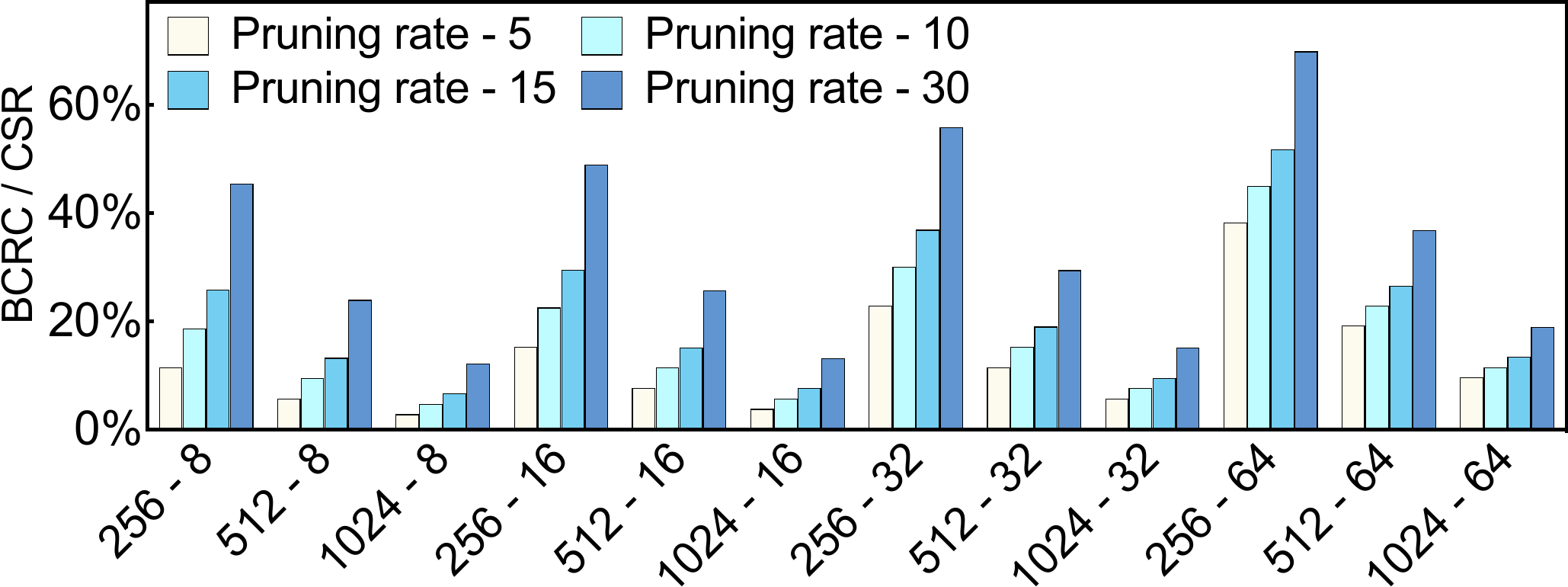}
    \caption{{\bf Extra data overhead comparison: BCRC/CSR with varied matrix sizes (x-axis) and pruning rates.}}
    \label{fig:eva_compact_storage}
\end{figure}

\begin{figure}[t]
    \centering
        \subfloat[Snapdragon 845]{
            \includegraphics[width=0.23\textwidth]{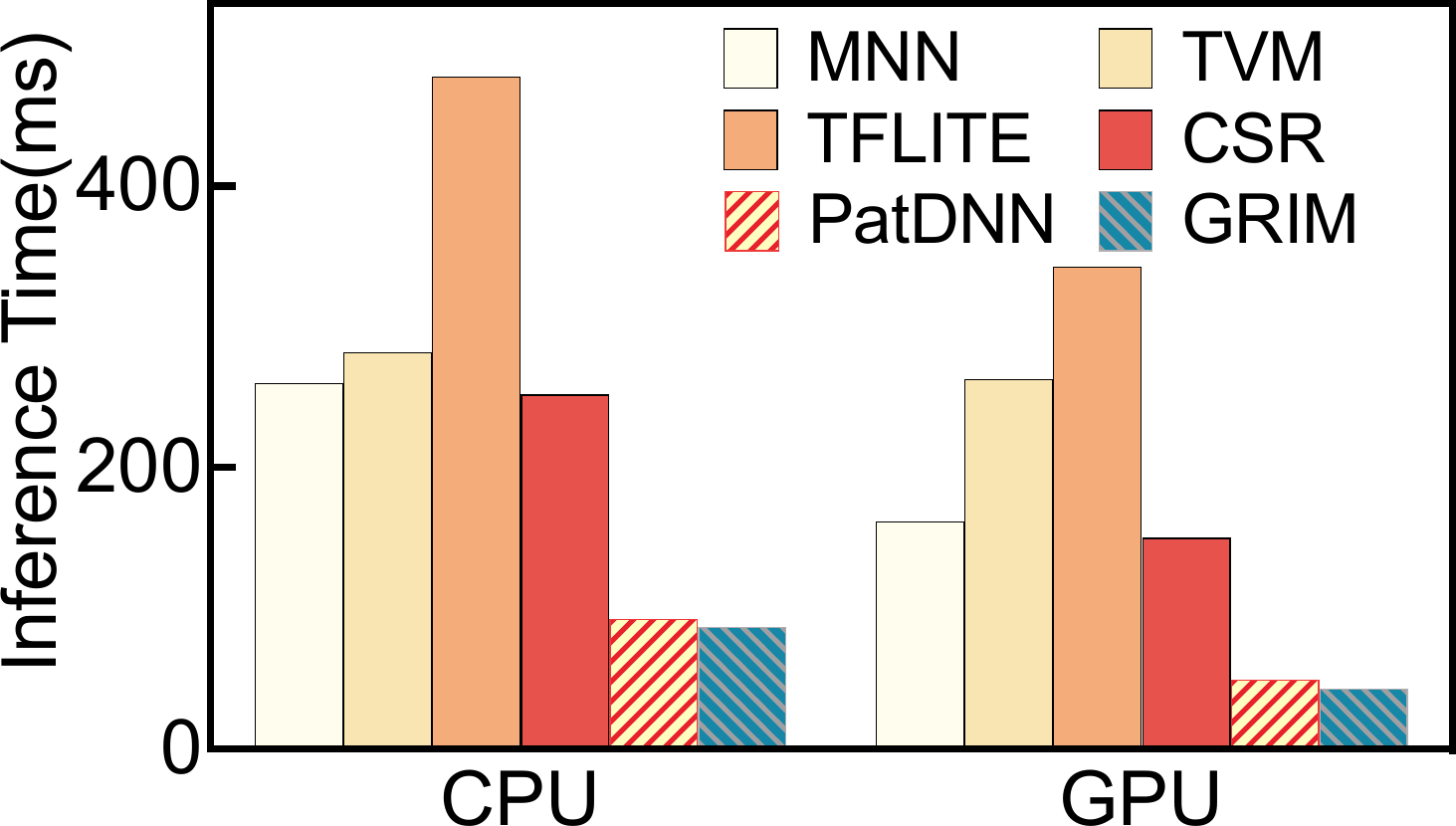}
        }
        \subfloat[Kirin 980]{
            \includegraphics[width=0.23\textwidth]{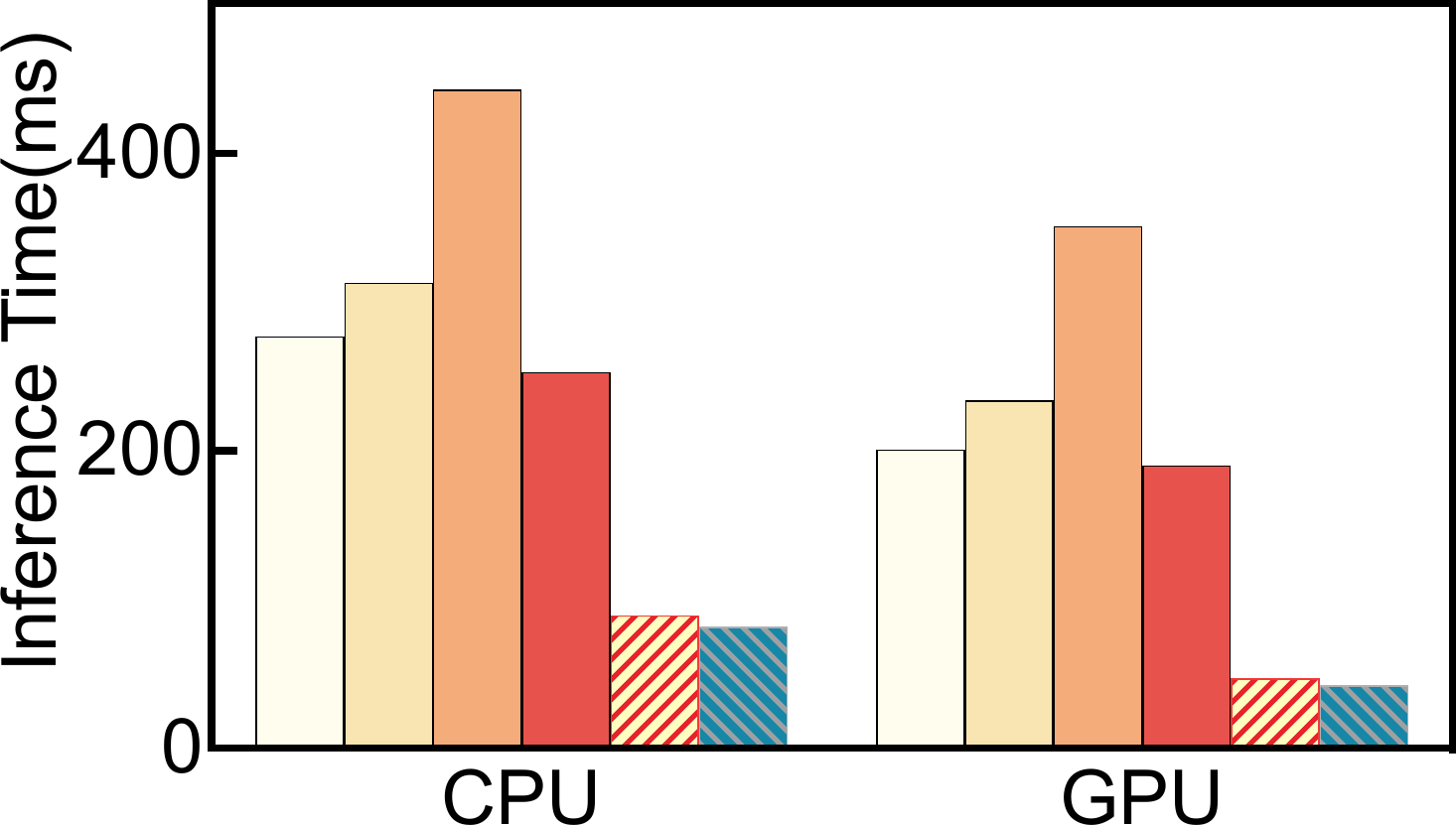}
        }
        \caption{{\bf Portability evaluation with VGG on ImageNet.}}
    \label{fig:eva_portability}
\end{figure}

\subsection{Performance Optimizations Break-down}

\projectname's superior performance mainly comes from two major resources. First, it has a fully optimized dense baseline, which is already $1.1\times$ to $1.6\times$ faster than TVM and MNN (extra optimizations are shown in Table~\ref{tab:dnn-frameworks}). Second, the flexible BCR pruning compresses the overall computation by $4\times$ to $20\times$. However, this computation reduction cannot transform to performance gains directly without further compiler optimizations due to the computation and memory access irregularity. This can be proved by CSR's performance in Figure~\ref{fig:eva_overview_performance}. Because CSR cannot leverage our compiler optimization, although its computation is almost equivalent to \projectname's, a significant performance gap still exists between CSR and \projectname. 
This part carefully studies the impact of \projectname's compiler optimizations. Please notice that these optimizations are {\em only enabled by BCR pruning}. 

\begin{table}[!t]
\renewcommand{\arraystretch}{1.2}
\vspace{1.5em}
\caption{\bf VGG CONV layers characteristics.}
\label{tab:eva_vgg_shortname}
\centering
\includegraphics[width = 0.98\linewidth]{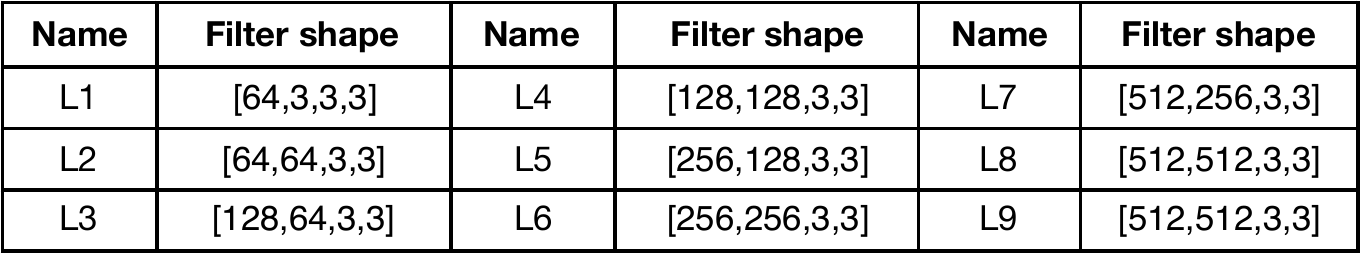}
        \vspace{-1.5em}
\end{table}

Figure~\ref{fig:eva_optimization} shows the performance improvement given by each optimization for VGG (on ImageNet)\footnote{RNN and other CNN results are omitted, very similar to VGG.}. The x-axis denotes the layers in VGG, and more detailed information is shown in Table~\ref{tab:eva_vgg_shortname}.  
This result uses the DNN execution code on BCR pruned models without any optimization ({\tt No-Opt}) as the evaluation baseline.
On CPU, matrix reorder ({\tt Reorder}) brings 1.21$\times$ to 1.88$\times$ speedup, register-level load redundancy elimination brings extra 1.11$\times$ to 3.51$\times$ speedup, and auto-tuning brings additional 0.31$\times$ to 1.45$\times$ speedup.
On GPU, these numbers are 1.30$\times$ to 2.88$\times$, 0.89$\times$ to 1.90$\times$, and 0.19$\times$ to 2.28$\times$, respectively. Matrix reorder optimization yields more benefits on GPU than CPU, because GPU has more threads and hence is more sensitive to thread divergence and load imbalance. We next characterize matrix reorder, load redundancy elimination, and compact storage optimizations to explain why they work. Auto-tuning and other optimizations are not further explained because their effects are more straightforward.

\noindent{\bf Effect of Matrix Reorder.} Figure~\ref{fig:eva_matrix_reorder} shows the number of non-zero weights (nnz) in each row for an RNN FC layer and a CNN CONV layer, respectively. Only the first 256 rows are plotted for readability. The nnz distribution is very random before matrix reorder ({\tt No-Reorder}), incurring significant thread divergence and load imbalance if these rows are processed by different threads. This distribution becomes much more regular after reorder ({\tt Reorder}). The rows with similar nnzs can be grouped together and each group can be processed by all threads simultaneously to minimize thread divergence and load imbalance.

\noindent{\bf Effect of LRE.} Figure~\ref{fig:eva_lre_count} reports the register load counts before and after the load redundancy elimination for multiple layers with different matrix sizes from both GRU (RNN) and VGG (CNN). It shows the number of register loads is significantly reduced with LRE optimization. This explains why LRE yields so obvious performance gains even after the traditional data locality optimizations like tiling.


\noindent{\bf BCRC VS CSR.}
Figure~\ref{fig:eva_compact_storage} shows the extra data storage overhead (i.e., the data size other than non-zero weights) for both BCRC and CSR with varied matrix sizes and pruning rates. BCRC saves 61.7\% to 97.1\%, 54.9\% to 95.2\%, 48.3\% to 93.3\%, and 30.1\% to 87.7\%  extra data over CSR for different pruning rates. This results in up to 48.5\%, 47.6\%, 46.6\%, and 43.8\% overall data reduction.

\subsection{Portability Evaluation}

We also run \projectname on two other  phones to validate its portability. We got very similar performance comparison results as above, which are omitted due to the space limitation. Figure~\ref{fig:eva_portability} reports the performance comparison of VGG (the most complex/largest DNN in our evaluation) between \projectname and other frameworks. On both platforms, \projectname outperforms others for both CPU and GPU, proving \projectname's good performance portability.
 \projectname's design and optimization are general, not specific to any brand or type of mobile device. \projectname is also less sensitive to resource constraints because of its high pruning rate, so its performance is stable on other mobile devices with even weaker computation power and smaller memory capabilities (e.g., Raspberry Pi).

\section{Related Work}\label{sec:related}



There have been many efforts on DNN acceleration frameworks on mobiles, like
DeepEar~\cite{lane2015deepear},
MCDNN~\cite{han2016mcdnn},
DeepMon~\cite{huynh2017deepmon}, 
DeepSense~\cite{yao2017deepsense},
DeepCache~\cite{xu2018deepcache},  etc.
Among those
TFLite~\cite{TensorFlow-Lite},
TVM~\cite{chen2018tvm},
Alibaba MNN~\cite{Ali-MNN}, 
and PatDNN \cite{niu2020patdnn}
are four state-of-the-art end-to-end acceleration frameworks for mobiles. 
Please note that those four do not or only partially support RNNs.
Also, these prior work do not utilize sparse DNN model inference, except for PatDNN.
Table~\ref{tab:dnn-frameworks} compares the major optimizations in TFLite, TVM, and MNN with \projectname. 
Basically, TFLite, TVM, and MNN are optimized for dense DNNs.


Table~\ref{tab:dnn-pruning} compares \projectname with PatDNN~\cite{niu2020patdnn}, an effort  sharing the most similarities with \projectname.
PatDNN does not support RNNs, because the pattern-based pruning it uses results in a fine-grained structured sparsity that only applies to weight tensors of the CONV layers.
So next we will further explain the differences between them on CNNs.
The fine-grained structured sparsity scheme by our BCR pruning achieves a higher pruning rate than PatDNN without accuracy loss.
Because (i) our fine-grained structured sparsity scheme has higher flexibility since PatDNN performs pruning only within CONV filter kernels, while our BCR pruning is performed within in a desirable block size, which does not necessarily follow the CONV filter kernels; (ii) GRIM uses a systematic optimization approach to determine the hyperparameters for pruning, while PatDNN uses a heuristic approach to select pattern candidates.
From the compiler optimizations, our GRIM introduces a new Domain Specific Language to offer users more flexibility of using existing DNNs or creating new ones, thus improving the programmability and productivity in DNN programming. Furthermore, our compiler optimizations are fully customized for our BCR pruning techniques.
Lastly, PatDNN and GRIM are complementary to each other and can be combined.

\begin{table}[t]
\scriptsize
\vspace{2.0em}
\caption{\bf DNN acceleration frameworks on mobile.}\label{tab:dnn-frameworks}
\vspace{-1.5em}
{\setlength{\tabcolsep}{5.0pt}
\begin{tabular}{llcccc}\\
\toprule
DNNs
 & Optimization Knobs & TFLite   & TVM  & MNN & {\bf GRIM}\\
\midrule
& Parameters auto-tuning & N & Y & N & {\bf Y}\\
& CPU/GPU support & Y & Y & Y & {\bf Y}\\
Dense & Half-floating support & Y & Y & Y & {\bf Y} \\
& Computation graph opt. & Y\tmark[$!$] & Y\tmark[*] & Y\tmark[$!$] & {\bf Y}\tmark[**] \\
& Tensor optimization & Y\tmark[$!$] & Y\tmark[$\dag$] & Y\tmark[$!$] & {\bf Y}\tmark[$\dag\dag$]\\
& RNN opt support & Y\tmark[$p$] & N & N & {\bf Y}\\
\midrule

& Sparse DNN model support & N & N & N & {\bf Y}\\
& BCR pruning & N & N & N & {\bf Y}\\
Sparse 
& Matrix reordering & N & N & N & {\bf Y}\\
& Opt. sparse kernel code gen & N & N & N & {\bf Y}\\
& Auto-tuning sparse models & N & N & N & {\bf Y}\\
\bottomrule
\multicolumn{5}{l}{{\tiny * Operator fusion, constant folding, static memory plan, data layout transform}} \\
\multicolumn{5}{l}{{\tiny ** Besides above in *, operation replacement}} \\
\multicolumn{5}{l}{{\tiny $\dag$ Scheduling, nested parallelism, tensorization, explicit memory latency hiding}} \\
\multicolumn{5}{l}{{\tiny $\dag\dag$ Besides above in $\dag$, dense kernel reordering, SIMD operation optimization}} \\
\multicolumn{5}{l}{{\tiny $!$ Similar optimizations as TVM, but less advanced}}\\
\multicolumn{5}{l}{{\tiny $p$ Latest version (partially) supports some RNN executions w/o RNN-specific opts.}}
\end{tabular}
}
\end{table} 

\begin{table}[t]
\scriptsize
\vspace{1.5em}
\caption{\bf Comparison between \projectname and PatDNN.}\label{tab:dnn-pruning}
\vspace{-1.5em}
{\setlength{\tabcolsep}{3.0pt}
\begin{tabular}{llcccc}\\
\toprule
Approach & Generality & Granularity & Pruning rate   & Prune Strategy & DSL \\
\midrule
PatDNN  & CNN    & Fine-grain  & Low & Empirical & N\\
\midrule
\projectname   & CNN\&RNN & Fine-grain  & High  & Systematical & Y\\
\bottomrule
\end{tabular}
}\vspace{-4.0ex}
\end{table}

    

{There are some other efforts that explore model compression to accelerate DNN executions including \cite{liu2015sparse}, DeftNN~\cite{hill2017deftnn}, SCNN~\cite{parashar2017scnn}, and AdaDeep~\cite{liu2018demand}. However, they either require new hardware support, or need a trade-off between inference acceleration performance and accuracy, or do not target mobile platforms.} Their focuses are different from \projectname.
\section{Conclusion}\label{sec:conclusion}


This paper presents a novel mobile inference acceleration framework GRIM that is general to both CNNs and RNNs and that achieves real-time performance and high accuracy, leveraging fine-grained structured sparse model inference and compiler optimizations for mobile devices. We begin with design of a new fine-grained structured sparsity scheme through the BCR pruning techniques. 
Our GRIM framework consists of two parts: (a) the compiler optimizations of execution code generation for real-time mobile inference; and (b) the BCR pruning optimizations for determining pruning hyperparameters and performing weight pruning. For CNNs, we compare with Alibaba MNN, TVM, TensorFlow-Lite, a sparse DNN inference implementation based on CSR format, and PatDNN, achieving significant speedups in the end-to-end inference time. For RNNs, we compare with ESE. GRIM achieves comparable end-to-end inference time with significantly higher energy efficiency.
\textcolor{revision}{
\projectname also has the potential to support other neural networks (e.g, transformers). 
The extremely deep nature of transformer models (e.g., BERT and its variants, and GPT, etc) requires more careful designs and optimizations of \projectname. This serves as a promising direction of our future work.
Particularly, our on-going study shows that \projectname can achieve a $1.8\times$ pruning rate on BERT-base models without notable accuracy loss.
}

\section*{Acknowledgements}

This   research   is   partially   funded   by   National   Science Foundation CCF-2047516 (CAREER), and CCF-1901378 and CCF-1937500,  Army  Research  Office/Army  Research  Laboratory  via  grant  W911NF-20-1-0167  (YIP)  to  Northeastern University, and Jeffress Trust Awards in Interdisciplinary Research to William \& Mary.
 Any opinions, findings, and conclusions or recommendations expressed in this material are those of the authors and do not necessarily reflect the views of NSF, Army  Research  Office, or Thomas F. and Kate Miller Jeffress Memorial Trust.




\bibliographystyle{IEEEtran}
\bibliography{IEEEabrv, ref_new.bib}

\begin{IEEEbiography}[{\includegraphics[width=1in,height=1.25in,clip,keepaspectratio]{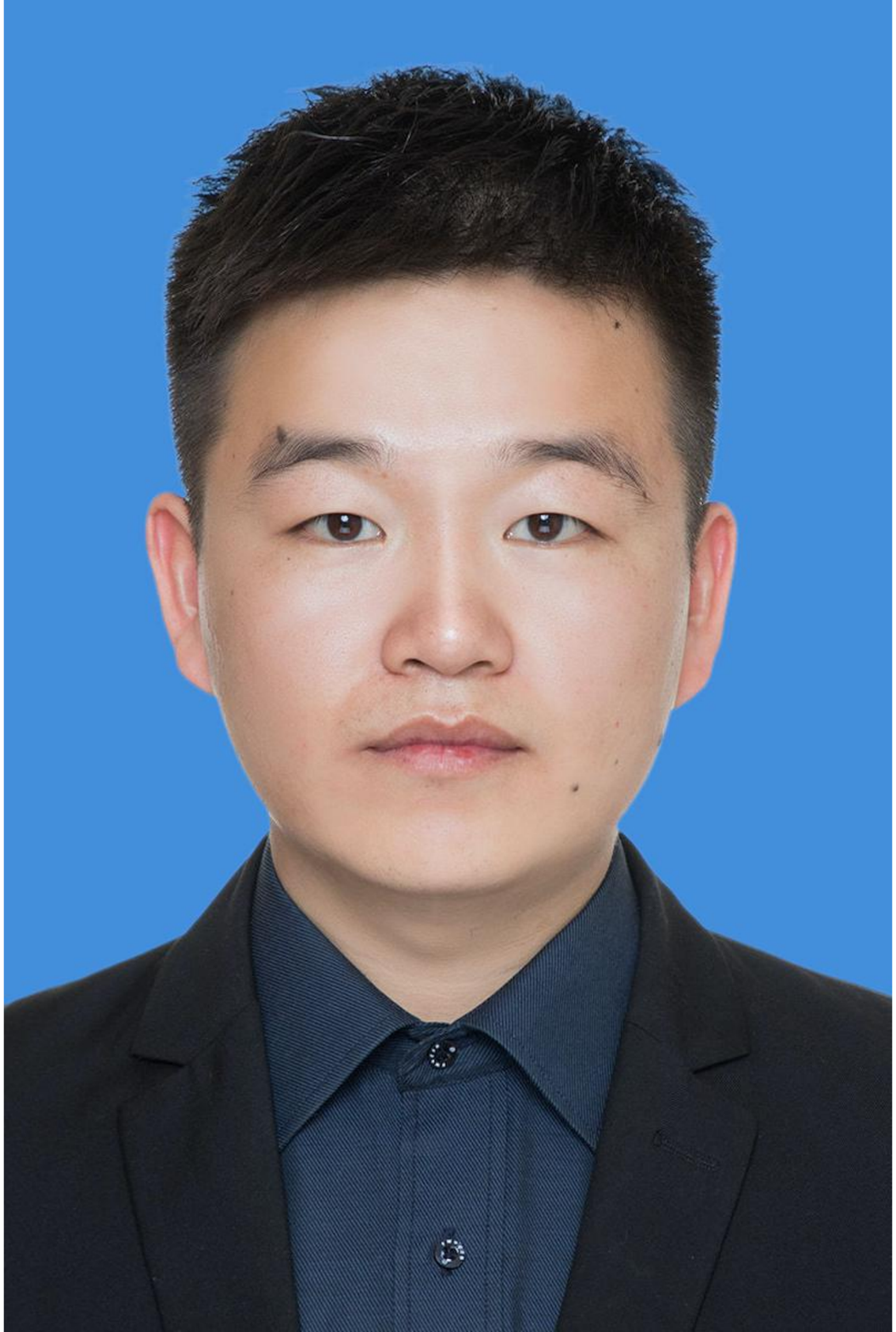}}]{Wei Niu}
received the B.E. degree in software engineering from Beihang University, Beijing, China, in 2016. He is currently pursuing the Ph.D. degree under Professor Bin Ren in computer science, William \& Mary, Williamsburg, VA, U.S. His current research interests include compiler optimizations for model compression of deep neural networks and computer architecture \& heterogeneous computing and machine Learning \& deep Learning.
\end{IEEEbiography}

\begin{IEEEbiography}[{\includegraphics[width=1in,height=1.25in,clip,keepaspectratio]{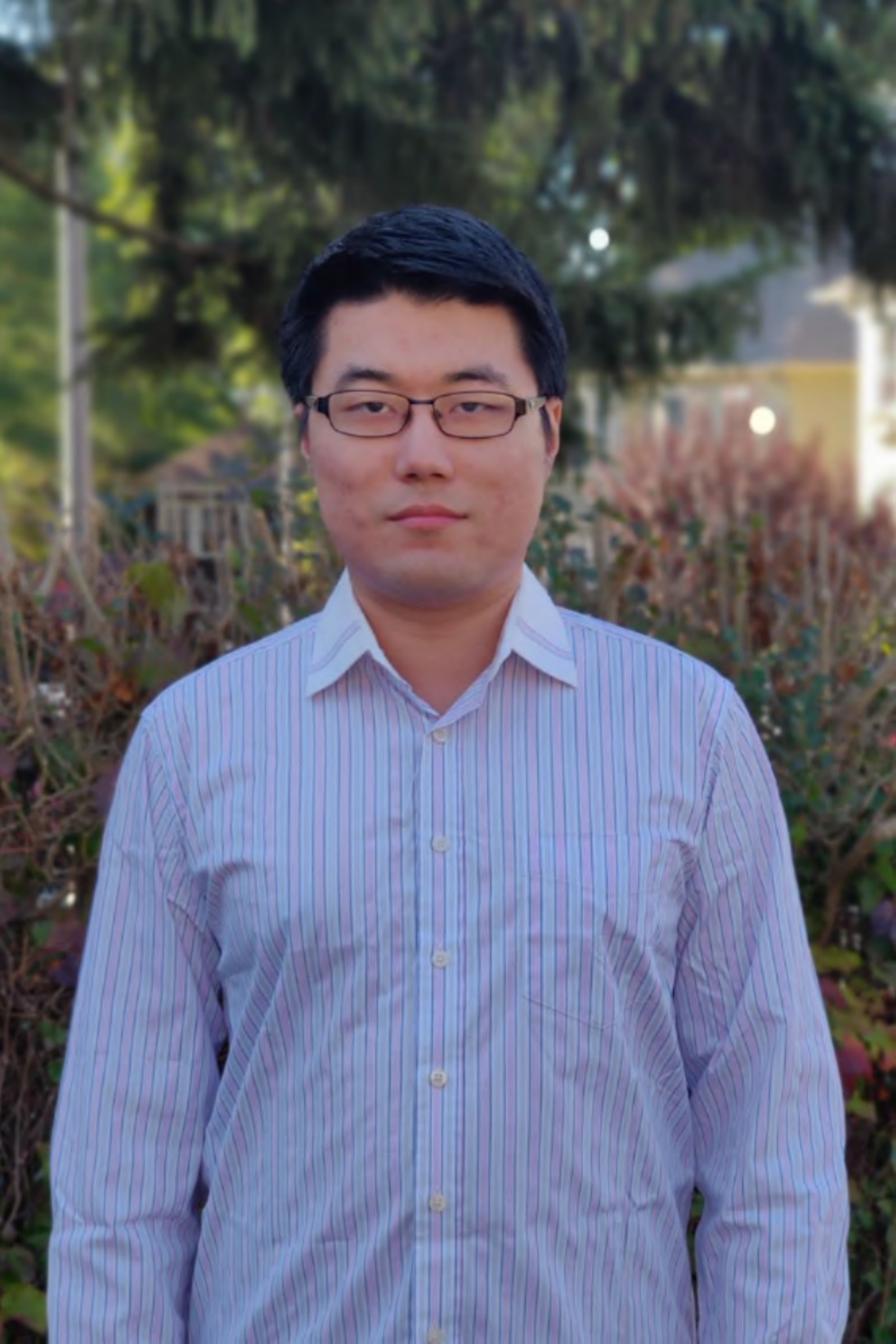}}]{Zhengang Li}
received the B.E. degree in electronic information engineering from Zhejiang University, Zhejiang, China, in 2017. He is currently pursuing the Ph.D. degree under Professor Yanzhi Wang in computer engineering, Northeastern University, Boston, MA, U.S. His current research interests include  model compression of deep neural networks, machine learning algorithm and computer vision.
\end{IEEEbiography}

\begin{IEEEbiography}[{\includegraphics[width=1in,height=1.25in,clip,keepaspectratio]{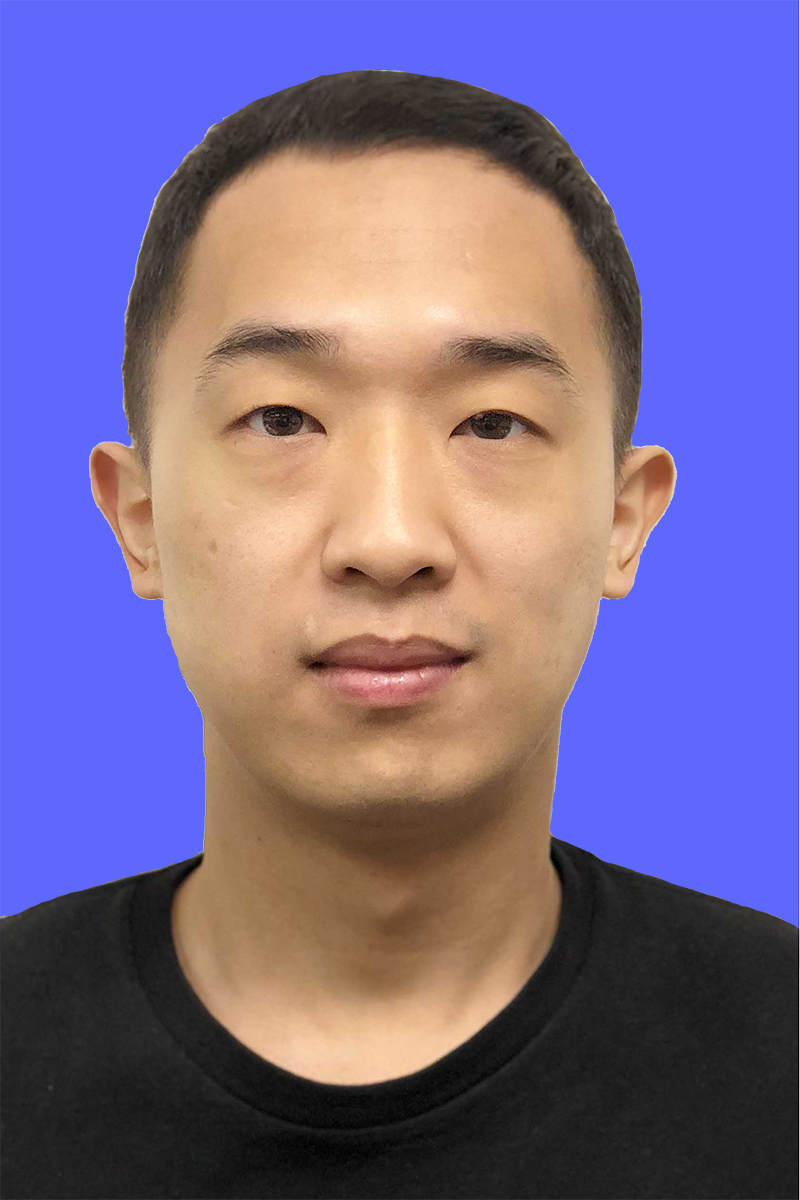}}]{Xiaolong Ma} 
received the MS degree in Electrical Engineering in Syracuse University, Syracuse, U.S. He is currently a Ph.D. candidate in the Department of Electrical and Computer Engineering in Northeastern University (NEU), Boston, U.S., supervised by Professor Yanzhi Wang. His research interests include machine learning algorithm, high performance computing and computer vision.

\end{IEEEbiography}

\begin{IEEEbiography}[{\includegraphics[width=1in,height=1.25in,clip,keepaspectratio]{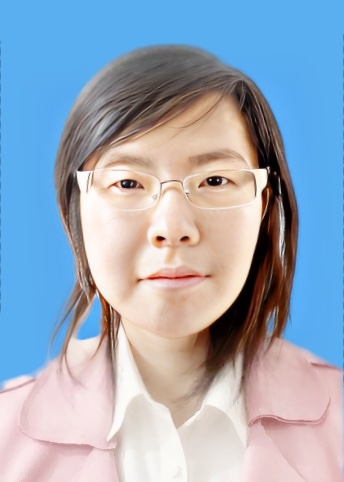}}]{Peiyan Dong}
received the B.E. degree in electronic and information engineering from  Beijing Institute of Technology, Beijing, China, in 2016. She is currently pursuing the Ph.D. degree under Professor Yanzhi Wang in computer engineering, Northeastern University, Boston, MA, U.S. Her current research interests include Model Compression for Deep learning \& Speech Recognition and Synthesis \& Deep Learning.
\end{IEEEbiography}

\begin{IEEEbiography}[{\includegraphics[width=1in,height=1.25in,clip,keepaspectratio]{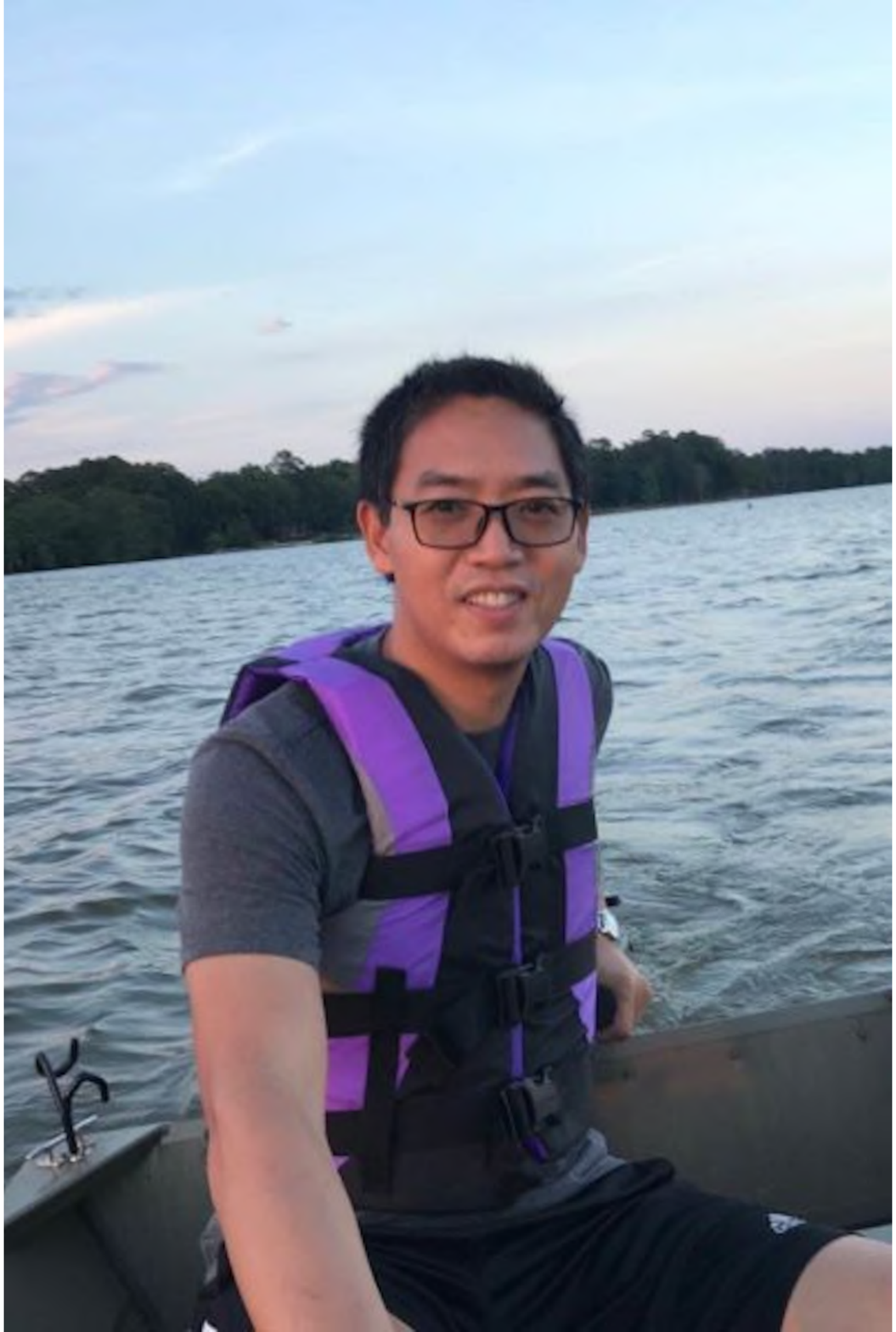}}]{Dr. Gang Zhou}
is currently an Associate Professor in the Computer Science Department at William \& Mary. He was Graduate Program Director from 2015 to 2017. He received his Ph.D. degree from the University of Virginia in 2007 under Professor John A. Stankovic. He has published more than 100 papers in the areas of wireless networks, sensor systems, internet of things, smart health, mobile and ubiquitous computing. There are over 7300 citations of his papers per Google Scholar. He serves or served on the Journal Editorial Board of (1) ACM Transactions on Sensor Networks, (2) IEEE Internet of Things, (3) Elsevier Computer Networks, and (4) Elsevier Smart Health. He served as NSF, NIH, and GENI proposal review panelists multiple times. He also received an award for his outstanding service to the IEEE Instrumentation and Measurement Society in 2008. He is a recipient of the Best Paper Award of IEEE ICNP 2010. He received NSF CAREER Award in 2013. He received a 2015 Plumeri Award for Faculty Excellence. He is a Senior Member of IEEE and ACM.
\end{IEEEbiography}

\begin{IEEEbiography}[{\includegraphics[width=1in,height=1.25in,clip,keepaspectratio]{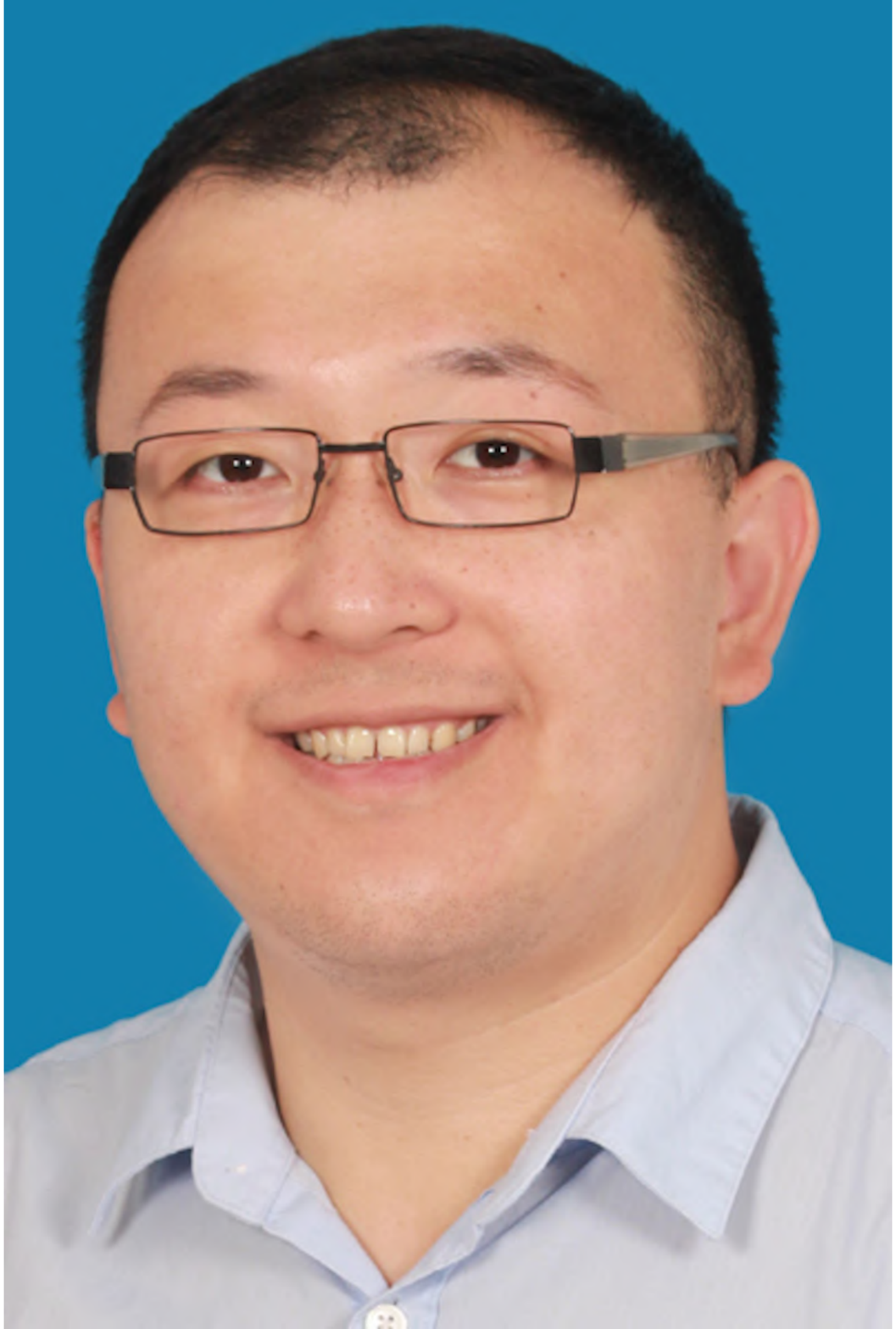}}]{Xuehai Qian}
received the BSc degree from Beihang University, China, in 2004, the MSc degree from the Institute of Computing Technology, Chinese Academy of Sciences, China, in 2007, and the PhD degree from the Computer Science Department, University of Illinois at Urbana-Champaign, in 2013. Currently, he is an assistant professor with the Ming Hsieh Department of Electrical Engineering and the Department of Computer Science, University of Southern California. His research interests include the fields of computer architecture, architectural support for programming productivity, and correctness of parallel programs.
\end{IEEEbiography}

\begin{IEEEbiography}[{\includegraphics[width=1in,height=1.25in,clip,keepaspectratio]{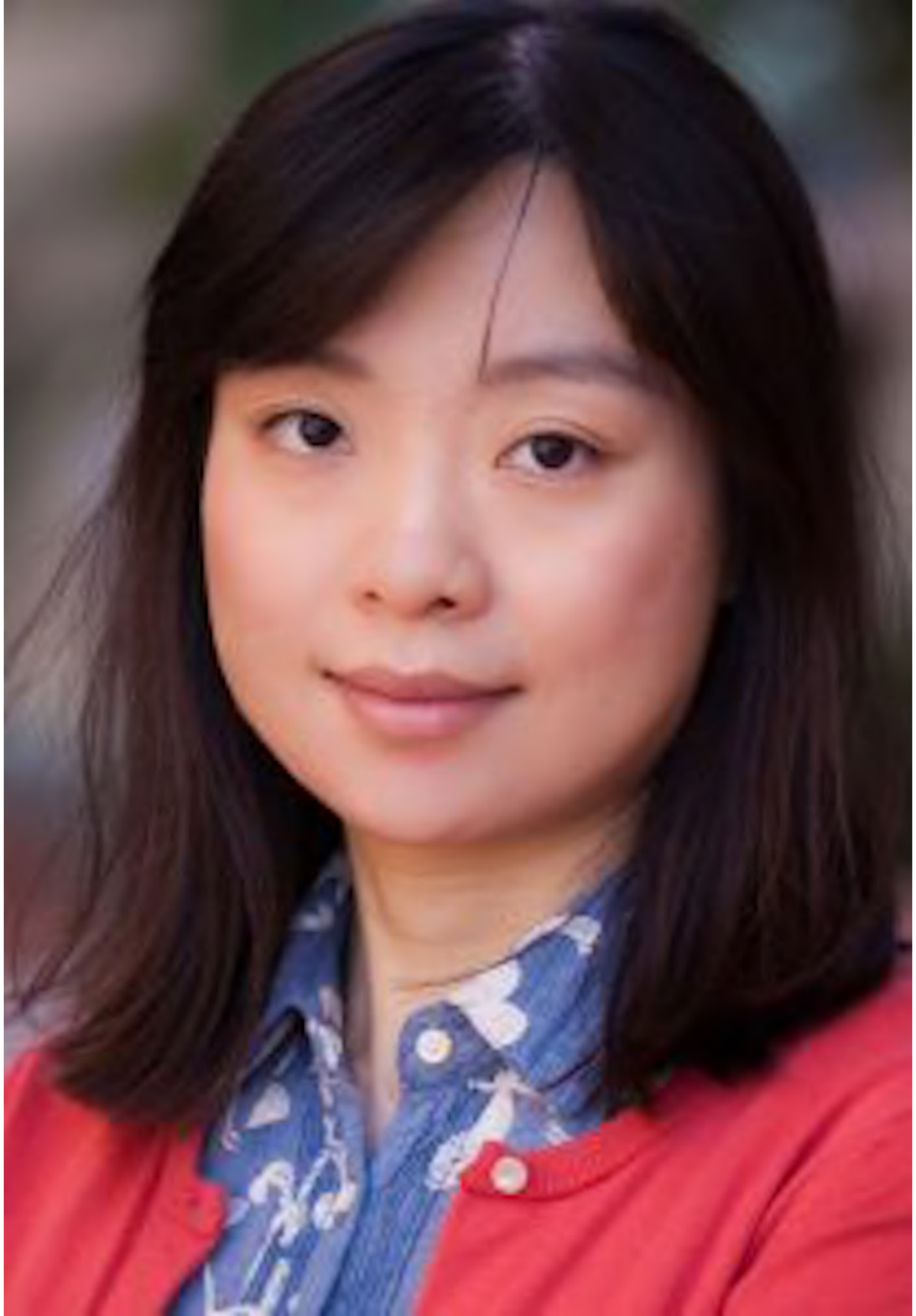}}]{Xue Lin (S’12–M’16)}
received the B.S. degree from Tsinghua University, Beijing, China, in 2009, and the Ph.D. degree in electrical engineering from the University of Southern California, Los Angeles, CA, USA, in 2016, under the supervision of Prof. M. Pedram. She is currently an Assistant Professor with the Department of Electrical and Computer Engineering, Northeastern University, Boston, MA, USA. Her current research interests include high-performance computing and mobile cloud computing systems, near-threshold computing for low-power embedded systems, and machine learning and computing in cyber-physical systems. She has published about 50 papers in the above areas.
\end{IEEEbiography}

\begin{IEEEbiography}[{\includegraphics[width=1in,height=1.25in,clip,keepaspectratio]{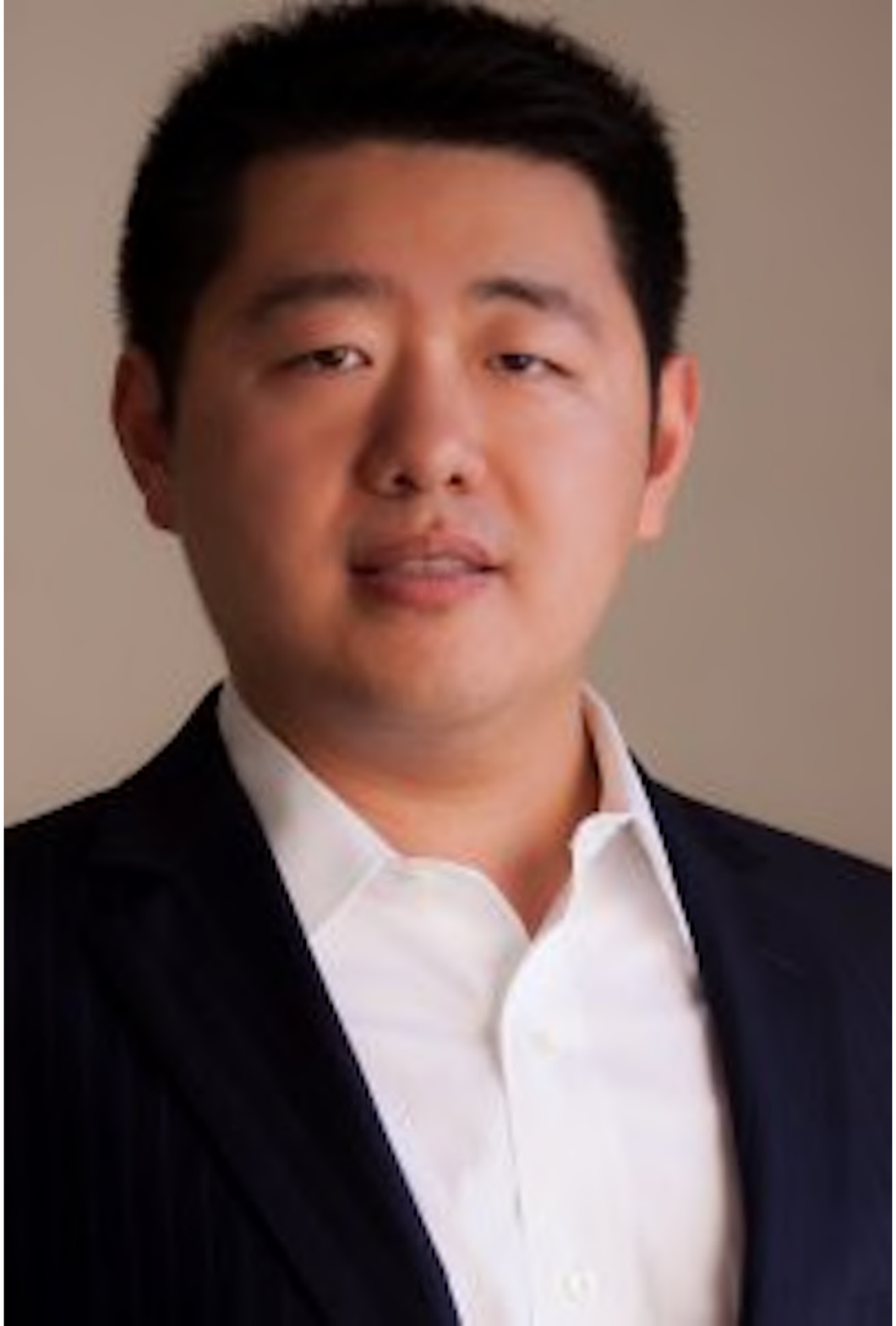}}]{Yanzhi Wang (Member, IEEE)}
received the B.S. degree in electronic engineering from Tsinghua University, Beijing, China, in 2009, and the Ph.D. degree in computer engineering from the University of Southern California, Los Angeles, CA, USA, in 2014. His research interests include energy-efficient and high-performance implementations of deep learning and artificial intelligence systems, emerging deep learning algorithms/systems, generative adversarial networks, and deep reinforcement learning.
\end{IEEEbiography}

\begin{IEEEbiography}[{\includegraphics[width=1in,height=1.25in,clip,keepaspectratio]{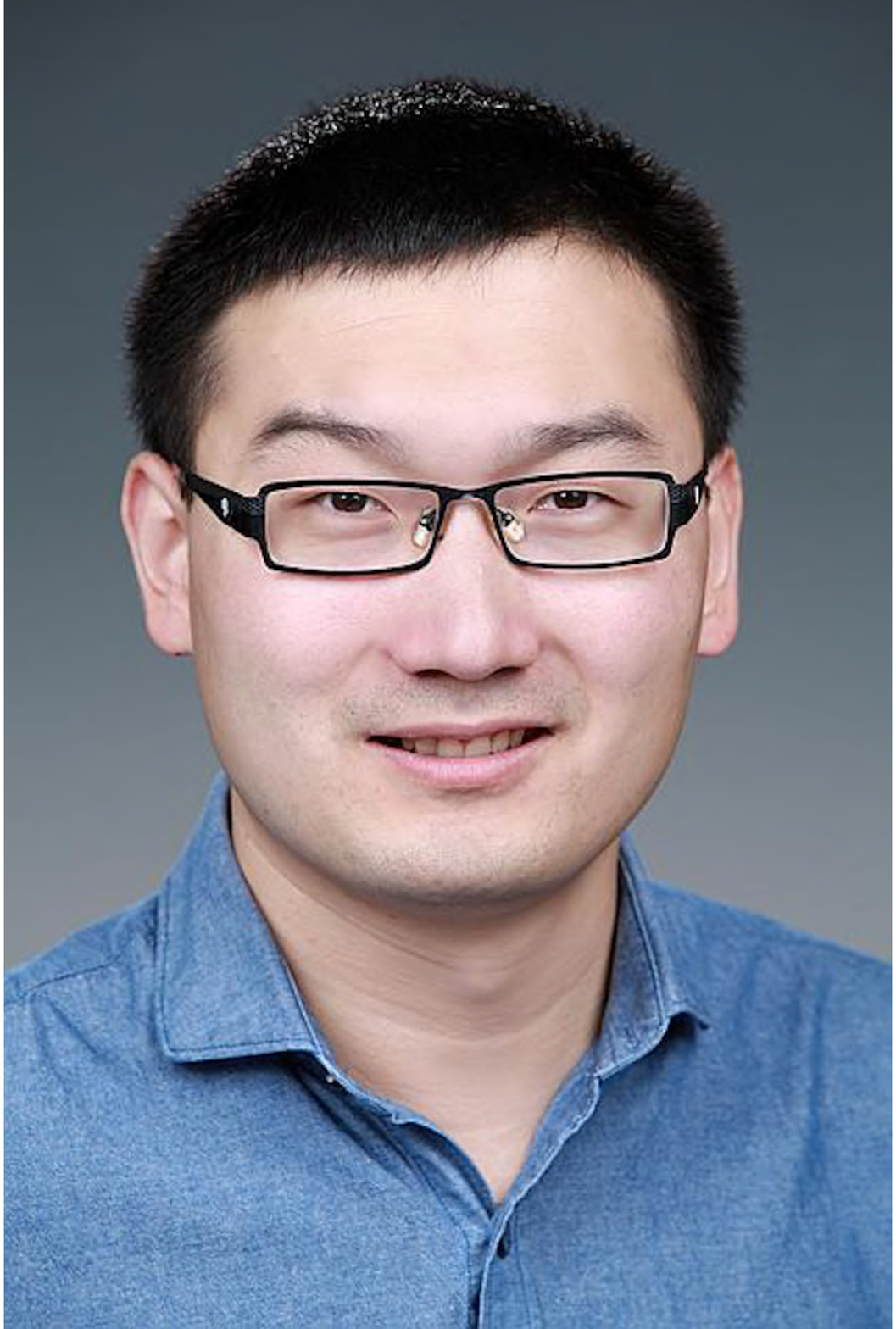}}]{Bin Ren}
received the B.S. degree from Beihang University, China, in 2006, and the M.S. and Ph.D. degrees from the Department of Computer Science and Engineering, Ohio State University, Columbus, Ohio, in 2013 and 2014, respectively. He was a post-doctoral research associate with the High-Performance Computing Group, Pacific Northwest National Laboratory. He is currently an assistant professor with the Department of Computer Science, William \& Mary. His research interest includes software systems, specifically programming systems, compiler and programming language support for parallel computing, and high-performance machine learning/deep learning systems. 
\end{IEEEbiography}

\end{document}